\documentclass{article}

\usepackage[preprint]{neurips_2025}

\usepackage[utf8]{inputenc} % allow utf-8 input
\usepackage[T1]{fontenc}    % use 8-bit T1 fonts
\usepackage{hyperref}       % hyperlinks
\usepackage{url}            % simple URL typesetting
\usepackage{booktabs}       % professional-quality tables
\usepackage{amsfonts}       % blackboard math symbols
\usepackage{nicefrac}       % compact symbols for 1/2, etc.
\usepackage{microtype}      % microtypography
\usepackage{xcolor}         % colors
\usepackage{float}
\usepackage{multirow} 
\usepackage{makecell} 
\usepackage{graphicx} 
\usepackage{siunitx}
\usepackage{wrapfig}
\usepackage{amsmath} 
\usepackage{caption}
\usepackage{placeins}
\usepackage{algorithm}       
\usepackage{algpseudocode}  
\usepackage{enumitem}
\usepackage{tcolorbox}
\tcbuselibrary{listings,breakable}
\usepackage{amssymb}
\usepackage{subfig}
\title{Image Editing As Programs with Diffusion Models}
\author{
Yujia Hu, Songhua Liu, Zhenxiong Tan, Xingyi Yang, and Xinchao Wang\thanks{Corresponding Author} \\
National University of Singapore\\
\texttt{\{yujia.hu,songhua.liu,zhenxiong,xyang\}@u.nus.edu,xinchao@nus.edu.sg}
}

\begin{document}

\maketitle

\begin{figure}[h]
    \vspace{-0.6cm}
    \centering 
    \includegraphics[width=1.0\textwidth]{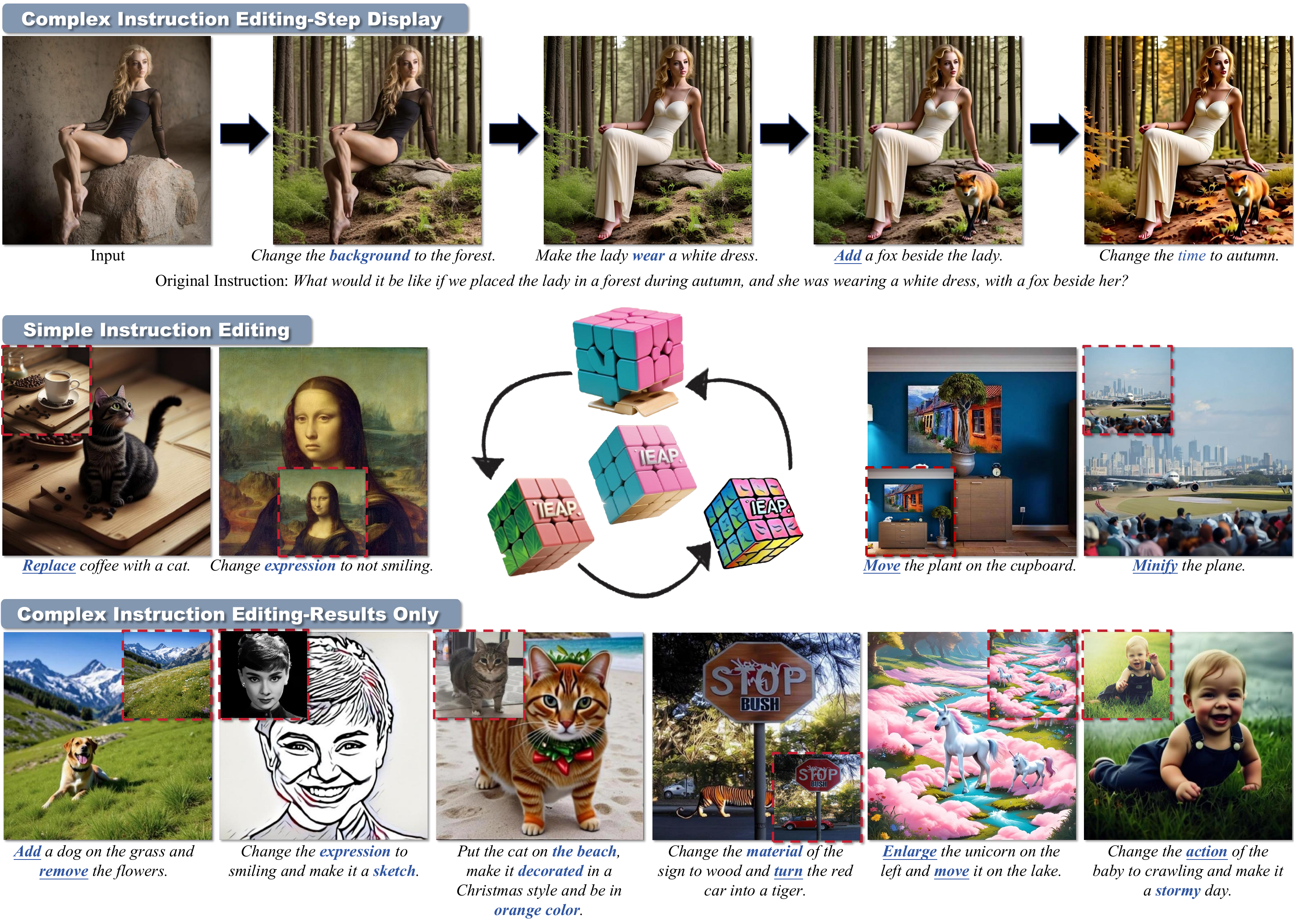}
    \caption{Visual results of our IEAP. Rows 1 and 3 showcase complex multi-step edits (Row 1 is further decomposed into individual instructions), while Row 2 shows single-instruction edits. Single instructions are \underline{underlined} if needing to be reduced to atomic operations.}
    \label{fig:teaser} 
    \vspace{-0.2cm}
\end{figure}

\begin{abstract}

While diffusion models have achieved remarkable success in text-to-image generation, they encounter significant challenges with instruction-driven image editing. 
Our research highlights a key challenge: these models particularly struggle with structurally inconsistent edits that involve substantial layout changes. 
To mitigate this gap, we introduce \textit{\textbf{I}mage \textbf{E}diting \textbf{A}s \textbf{P}rograms} (IEAP), a unified image editing framework built upon the Diffusion Transformer (DiT) architecture. 
At its core, IEAP approaches instructional editing through a \textbf{reductionist lens}, decomposing complex editing instructions into sequences of \textit{atomic} operations. 
Each operation is implemented via a lightweight adapter sharing the same DiT backbone and is specialized for a specific type of edit. 
Programmed by a vision-language model (VLM)-based agent, these operations collaboratively support arbitrary and structurally inconsistent transformations. 
By modularizing and sequencing edits in this way, IEAP generalizes robustly across a wide range of editing tasks, from simple adjustments to substantial structural changes. 
Extensive experiments demonstrate that IEAP significantly outperforms state-of-the-art methods on standard benchmarks across various editing scenarios. In these evaluations, our framework delivers superior accuracy and semantic fidelity, particularly for complex, multi-step instructions. Codes are available \href{https://github.com/YujiaHu1109/IEAP}{here}.

\end{abstract}

\section{Introduction}

% background of image editing
% difficulty: instructions are complex

Image editing lies at the heart of a wide range of applications from photo retouching and content creation to visual storytelling and scientific visualization \cite{oh2001image,barnes2009patchmatch,Wang_2018_CVPR}.
With the advent of diffusion models \cite{ho2020denoisingdiffusionprobabilisticmodels,rombach2022highresolutionimagesynthesislatent,podell2023sdxl}, the field has shifted towards highly precise and controllable manipulations \cite{parihar2024precisecontrol,epstein2023diffusion,tsagkas2024click}. The inherently progressive denoising process enables multi-stage pipelines \cite{ho2022cascaded,avrahami2022blended,avrahami2023blended} and localized editing methods \cite{couairon2022diffedit,zhang2023adding,shi2024dragdiffusion}, and its native support for multi-modal inputs has inspired unified frameworks that integrate heterogeneous signals within a single model \cite{li2023instructany2pix,fu2023guiding,yang2023uni,goel2024pair}.

More recently, text-to-image pipelines based on Diffusion Transformers (DiTs) \cite{peebles2023scalable,esser2024scaling,Flux2024} have set new standards in generative fidelity. However, their capacity for instruction-driven editing \cite{nguyen2024instruction,Huang_2025} remains under-explored.
Notably, although there are a few existing methods \cite{zhang2025incontexteditenablinginstructional,liu2025step1x} that have extended DiTs to instruction-driven editing, they are always restricted to a narrow set of common editing operations and lack evaluation on comprehensive editing tasks.

To address this limitation, we initiate a taxonomy study of image editing instructions to systematically assess the editing capabilities of current DiT-based conditional generation methods.
Our empirical analysis reveals an interesting performance dichotomy: While current methods demonstrate proficiency in structurally-consistent edits where the layouts of the input and output images remain aligned, they exhibit significant degradation when handling structurally-inconsistent operations that require layout modifications.

To overcome this issue, we introduce \textit{\textbf{I}mage \textbf{E}diting \textbf{A}s \textbf{P}rograms} (IEAP), a unified framework atop the DiT architecture which is capable of handling diverse types of editing operations efficiently and robustly in this paper.
Notably, we show that structurally-inconsistent instructions can in fact be reduced to a small set of simple operations, which are called as \textit{atomic} operations in our paper. 
Thus, instead of treating each edit as a monolithic, end‑to‑end task, IEAP levarages the Chain-of-Thought (CoT) reasoning \cite{wei2022chain} to break the original editing command into a sequence of atomic operations, which are namely Region of Interest (RoI) localization, RoI inpainting, RoI editing, RoI compositing and global transformation, and then executes them in a sequential manner via a neural program interpreter \cite{reed2015neural}.

The five atomic operations serve as the fundamental building blocks for complex editing tasks. 
As such, through the sequential combination of atomic operations, IEAP can robustly handle complex, multi‑step instructions that are typically confound in conventional end‑to‑end approaches. 

Extensive experiments show that our framework demonstrates state-of-the-art performance across standard benchmarks, excelling in both structural preservation and alteration tasks through atomic-level operation decomposition compared to other approaches.
Simultaneously, the CoT reasoning and programming pipeline of IEAP enable significantly more accurate and semantically more coherent edits under complex, multi-step instructions even compared to the leading proprietary models.

Our main contributions can be summarized as follows:
% \begin{itemize}
%     \item We introduce \textit{\textbf{I}mage \textbf{E}diting \textbf{A}s \textbf{P}rograms} (IEAP), a unified instruction‑driven editing framework that can deal with a vast spectrum of image editing instructions.
%     \item  
%     \item Experimental results demonstrate that xxx 
% \end{itemize}

\begin{itemize}
    \item We present a comprehensive taxonomy and empirical analysis of instruction-driven editing in DiT-based conditional generation, revealing a performance dichotomy between structurally-consistent and -inconsistent edits.
    % \item We propose a reductionist formulation that decomposes any editing instruction into five atomic operations, showing that layout-altering and complex edits can be expressed as programs over these primitives.
    \item We introduce \textit{\textbf{I}mage \textbf{E}diting \textbf{A}s \textbf{P}rograms} (IEAP), a unified framework on the DiT backbone that leverages CoT reasoning to parse free-form instructions into sequential atomic operations and then executes them sequentially by a neural program interpreter, thereby enabling robust handling of layout-altering and complex edits.
    % We introduce \textit{\textbf{I}mage \textbf{E}diting \textbf{A}s \textbf{P}rograms} (IEAP), a unified DiT-based framework that formulates the free-form editing instruction as a program over five atomic operations: using Chain-of-Thought reasoning to parse the instruction into a sequential graph of primitives and a neural program interpreter to execute them, thereby enabling robust handling of both simple and complex, layout-altering edits.
    \item Extensive experiments demonstrate that IEAP achieves state-of-the-art performance in both structure-preserving and -altering scenarios, delivering notably higher accuracy and semantic fidelity especially on complex, multi-step instructions compared to existing methods.
\end{itemize}

\section{Related Work}
\label{gen_inst}
\noindent \textbf{Conditional image generation.} Early conditional image generation approaches like ControlNet \cite{zhang2023adding} typically adopt plug-in control adapters to incorporate single condition \cite{avrahami2023spatext, gafni2022make,li2023gligen} like segmentation mask or diverse conditional inputs \cite{zhao2023uni,qin2023unicontrol,huang2023composer,mou2024t2i,xu2024prompt} to guide the generation of images.
% Recent advancements in DiT-based conditional generation have demonstrated significant progress. 
Recently, the field of conditional image generation has witnessed remarkable breakthroughs through the integration of DiTs \cite{esser2024scaling,peebles2023scalable,Flux2024}, with continuous innovations improving output quality and edit precision \cite{parihar2024precisecontrol}.
Some methods \cite{xiao2024omnigen,le2024one,xia2024dreamomni,chen2024unireal} aim to create a unified DiT foundation for versatile conditional image generation and editing by integrating diverse inputs within a single framework.
while approaches like OminiControl \cite{tan2024ominicontrol} and so on \cite{tan2025ominicontrol2,zhang2025easycontroladdingefficientflexible,mao2025aceinstructionbasedimagecreation,zhang2025incontexteditenablinginstructional,wu2025less} leverage LoRA-based fine-tuning \cite{hu2021loralowrankadaptationlarge} for lightweight and effective control.

\noindent \textbf{Instructional image editing.} Instruction-based image editing \cite{nguyen2024instruction,Huang_2025} enables intuitive, language-driven modifications of existing images. Early works like InstructPix2Pix \cite{brooks2023instructpix2pix} establishes paired instruction–image datasets for supervised fine-tuning of diffusion models. For subsequent works, some of them focus on architectural refinement \cite{mao2025aceinstructionbasedimagecreation,liu2025step1x,zhao2024ultraedit,li2023moecontroller,guo2024focus}, which introduce specialized conditioning units and multi-stage training to improve control granularity and consistency, others concentrate on data-centric enhancements \cite{zhang2023magicbrush,geng2024instructdiffusion,sheynin2024emu,chakrabarty2023learning}, that expand instruction coverage and diversify edit examples. 
% To better align outputs with human intent, methods like HIVE incorporate human feedback into the learning loop. 
Moreover, some approaches \cite{zhang2025nexus,huang2024smartedit,li2023instructany2pix,fu2023guiding}
has unified LLM-based \cite{openai2024gpt4technicalreport} language reasoning with diffusion-based synthesis in a single framework, and some \cite{yang2025textttcomplexeditcotlikeinstructiongeneration,zhang2024tierevolutionizingtextbasedimage} leverage CoT \cite{wei2022chain} and in-context learning \cite{gupta2022visualprogrammingcompositionalvisual} to enhance the reasoning ability of models for more complex editing tasks.
% while in-context and few-shot approaches such as ICEdit leverage demonstration examples at inference to adapt to novel editing commands. 
% an in-context editing framework for zero-shot instruction compliance using in-context prompting
More recently, some works \cite{feng2025dit4edit,zhang2025incontexteditenablinginstructional,liu2025step1x} have advanced image editing with DiTs. For instance, ICEdit \cite{zhang2025incontexteditenablinginstructional} leverages the in-context generation capabilities of large-scale DiTs to achieve flexible few-shot instruction editing, while Step1X-Edit \cite{liu2025step1x} focuses on large-scale data construction and multi-modal integration to enable general-purpose image editing with performance approaching proprietary models.

\section{Motivation}
\subsection{Preliminaries}
\label{sec:preliminaries}
% Introduce OminiControl

\noindent \textbf{Diffusion Transformer Fundamentals.} The image generation process of text-guided DiTs \cite{peebles2023scalable,esser2024scaling,Flux2024} is accomplished by successively denoising input tokens in multiple steps. At step $t$, the model processes: 
\begin{equation}
    \mathbf{S}_t = [\mathbf{X}_t, \mathbf{C}_T]
\end{equation}
where $\mathbf{X}_t \in \mathbb{R}^{N \times d}$ represents noisy image tokens and $\mathbf{C}_T \in \mathbb{R}^{M \times d}$ denotes text tokens, they share the embedding dimension $d$. Image tokens use Rotary Position Embedding (RoPE) \cite{su2024roformer} with spatial coordinates $(i,j)$, while text tokens fix positions at $(0,0)$, enabling Multi-Modal Attention (MMA) \cite{pan2020multi} mechanisms to model cross-modal interactions.

\noindent \textbf{Unified Conditioning Framework.} To integrate visual control signals, the prior work \cite{tan2024ominicontrol} extends the baseline formulation by incorporating encoded condition images:

\begin{equation}
    \mathbf{S}_t = [\mathbf{X}_t, \mathbf{C}_T, \mathbf{C}_I]
\end{equation}

where $\mathbf{C}_I \in \mathbb{R}^{N \times d}$ denotes latent tokens from condition images via the pretrained VAE encoder \cite{kingma2013auto,rombach2022high}. This unified sequence enables tri-modal fusion within transformer architectures, eliminating spatial misalignment inherent in feature concatenation baselines.

Moreover, an auxiliary adaptive positional encoding mechanism further preserves spatial consistency across these modalities by assigning coordinates to each token type with minimal overhead.

\noindent \textbf{Gap in Instruction‐Driven DiT Editing.} Despite the rapid advances in DiT-based conditional image generation \cite{tan2024ominicontrol,zhang2025easycontroladdingefficientflexible,mao2025aceinstructionbasedimagecreation,wu2025less}, research on instruction-driven editing \cite{nguyen2024instruction,Huang_2025} remains scarce. The few existing methods \cite{zhang2025incontexteditenablinginstructional,liu2025step1x} that do support instructional edits are typically confined to a small set of routine operations, and lack a comprehensive evaluation across diverse editing scenarios, leaving DiT’s true editing potential unclear. This gap motivates us to conduct a taxonomy study of DiT's ability in instructional image editing, which is detailed in Sec. \ref{sec:observation}.

\subsection{Preliminary Experiments and Observations}
\label{sec:observation}
\begin{figure}[t]
    \vspace{-0.3cm}
    \centering 
    \includegraphics[width=1.0\textwidth]{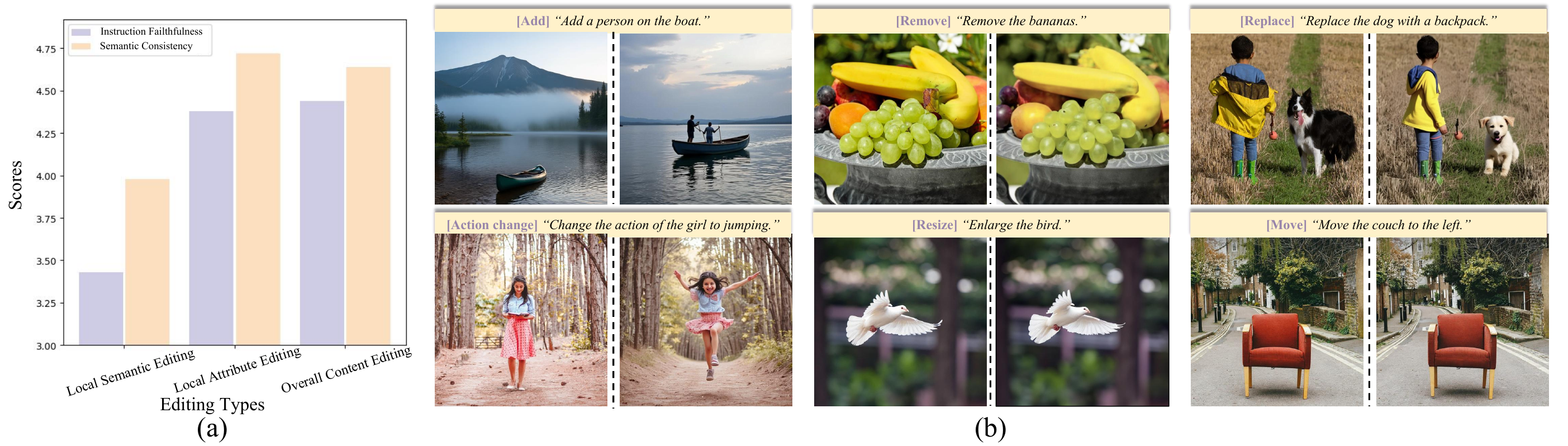}
    \vspace{-0.25cm}
    \caption{Results of our preliminary experiments. Figure (a) shows the GPT-4o scores for three editing types across instruction faithfulness and semantic consistency, ranging from 1 to 5. Figure (b) shows the representative failure cases from local semantic editing.}
    \label{fig:pre_exp_results} 
    \vspace{-0.25cm}
\end{figure}

To this end, we conduct a comprehensive evaluation of diffusion models for instruction‐driven editing, uncovering an interesting performance dichotomy: \textit{While these methods excel at structurally‐consistent edits, they falter dramatically on structurally‐inconsistent operations that demand explicit layout modifications.}

\noindent \textbf{Taxonomy and Experimental Setup.} To enable systematic analysis \cite{Huang_2025, yu2024anyedit,yang2025textttcomplexeditcotlikeinstructiongeneration}, we first categorize instruction-based image editing into three main types: local semantic editing, which modifies the identity, position or size, e.g., add, remove, replace, action change, move and resize; local attribute editing, which adjusts certain properties of objects, e.g., color change, texture change, appearance change, expression change, and background change; and overall content editing, which alters the whole image consistently, e.g., tone transfer and style change. 

Then we use AnyEdit dataset \cite{yu2024anyedit} and OminiControl \cite{tan2024ominicontrol} to train models on the above editing types, accompanied by GPT-4o \cite{openai2024gpt4ocard} to rate each edit on instruction faithfulness and semantic consistency.

\noindent \textbf{Results and Analysis.} As shown in Fig. \ref{fig:pre_exp_results}(a), both local attribute editing and overall content editing attain relatively high GPT-4o scores, whereas local semantic editing exhibits a notable performance drop.
As illustrated in Fig. \ref{fig:pre_exp_results}(b), the cases of ``add'' and ``action change'' alter unrelated areas like the background, and the remaining four cases demonstrate a complete failure.

We attribute this discrepancy to the fact that, unlike local attribute and overall content edits, local semantic edits require explicit spatial‐layout modifications. For instance, ``add'' and ``delete'' operations necessitate instance-level scene recomposition, while ``move'' and ``resize'' further demand precise coordinate system recalibration.

\noindent \textbf{Key Insight.} Based on the above analysis, 
spatial‐layout modification remains a critical challenge for diffusion‐based editing models; conversely, edits that preserve the original layout demonstrate substantially better performance. 
We speculate that, with limited training data, it is difficult for the model to learn the complex patterns underlying layout‐changing tasks. Although DiT architectures \cite{peebles2023scalable,esser2024scaling,Flux2024} employ powerful full‐attention mechanisms to capture long‐range dependencies, they still struggle with editing operations that require nontrivial scene reconfiguration. 

Due to the combinatorial complexity of spatial-layout modifications and the empirical limitations of DiT architectures, we propose to simplify the layout‐editing paradigm through decomposition, which is detailed in Sec. \ref{methods}.

\section{Methods}
\label{methods}

\subsection{Program with Atomic Operations}
\begin{figure}[t]
    \vspace{-0.3cm}
    \centering 
    \includegraphics[width=1.0\textwidth]{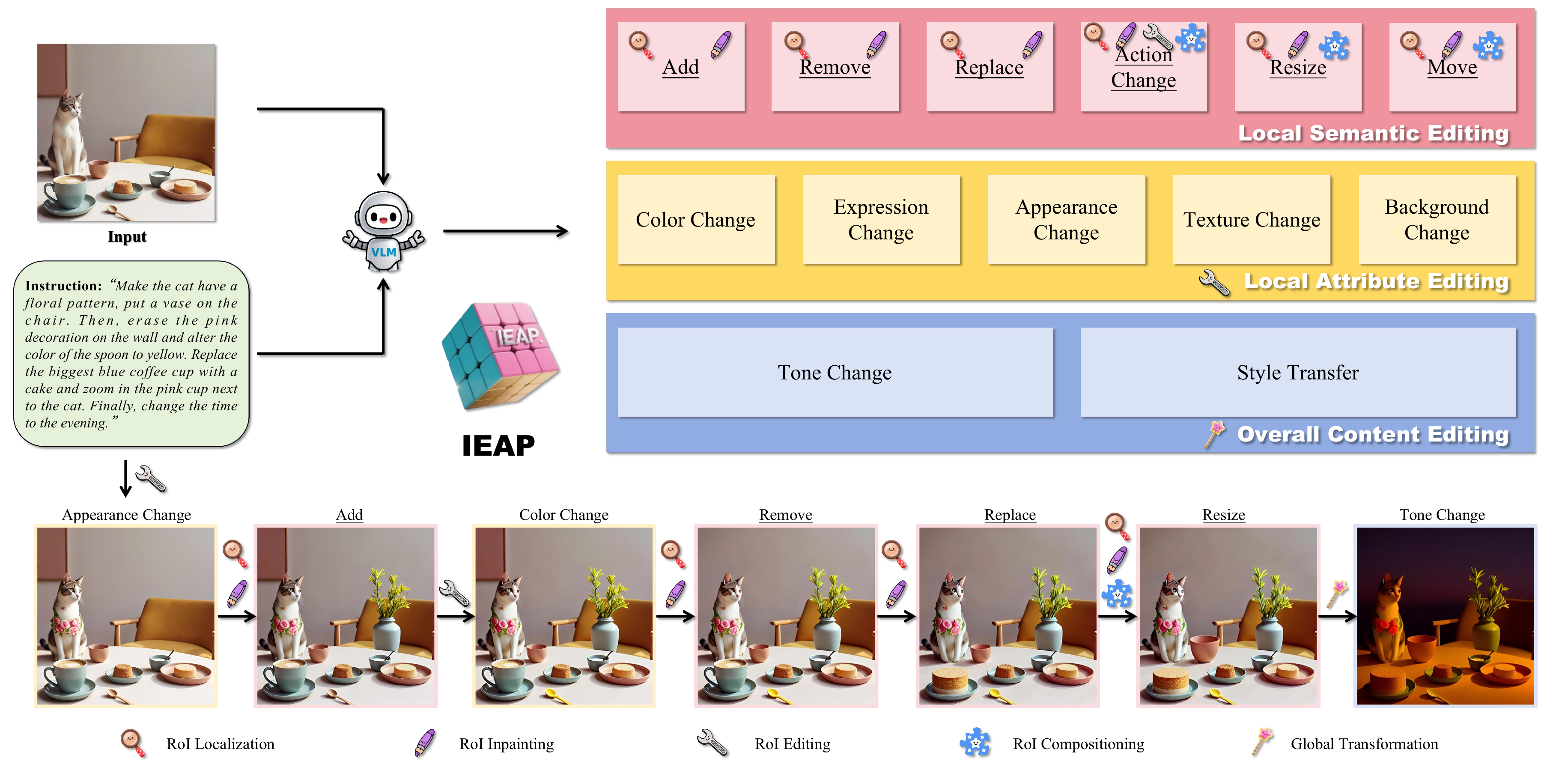}
    \caption{Our pipeline. The original instruction is first parsed by a VLM into atomic operations, which are then sequentially executed via a neural program interpreter.}
    \vspace{-0.3cm}
    \label{fig:pipeline} 
\end{figure}
The insight in Sec. \ref{sec:observation} motivates us to decouple semantic and spatial reasoning. Building on this foundation, we propose a programmatic reduction framework that systematically decomposes complex editing instructions into modular atomic operations. 
Specifically, we first formulate instruction-driven image editing as an executable program via Chain-of-Thought (CoT) reasoning \cite{wei2022chain}, and then use a neural program interpreter \cite{reed2015neural} to transcode the reasoning graph into a dynamic execution plan, sequentially invoking relevant atomic modules.

\subsection{General Pipeline}
% RoI Localization, RoI Inpainting, RoI Editing, RoI Compositing and Global Transformation. 
We abstract all editing instructions into five atomic primitives: 
(1) RoI Localization: Identify and isolate the relevant region in the image that the instruction refers to, serving as the spatial grounding step for subsequent localized edits;
(2) RoI Inpainting: Introduce new visual content or remove existing elements within the localized region, enabling semantic-level additions, substitutions, or deletions;
(3) RoI Editing: Modify visual attributes within the region, such as color, texture, or appearance, to reflect fine-grained property changes specified by the instruction;
(4) RoI Compositing: Reintegrate the edited region into the full image while preserving spatial coherence and visual continuity;
(5) Global Transformation: Adjust the overall content for coherent full-image modifications, such as changing the illumination, weather, or style of the whole image.
% Performs fulfill high-level instructions that require coherent modifications across the entire image.
% \yxy{example: XXX}

The overall pipeline is shown as Fig. \ref{fig:pipeline}. We reduce any editing instruction into an arbitrary combination of the five atomic operations described above, which can be formulated as:
\begin{equation}
{\textit{T}} \equiv \bigoplus_{k=1}^K \mathcal{A}_k, \quad \ 
\mathcal{A}_k \in \{ \mathcal{A}_{\text{loc}}, \mathcal{A}_{\text{inp}}, \mathcal{A}_{\text{edit}}, \mathcal{A}_{\text{comp}}, \mathcal{A}_{\text{global}} \}
\end{equation}
where $T$ denotes the free‐form editing instruction, $\bigoplus$ represents the sequential program combination, $K$ is the number of atomic operations, $\mathcal{A}_{\text{loc}}$, $\mathcal{A}_{\text{inp}}$, $\mathcal{A}_{\text{edit}}$, $\mathcal{A}_{\text{comp}}$, and $\mathcal{A}_{\text{global}}$ represent the five atomic primitives respectively.

% We now describe each atomic primitive in detail.

\noindent \textbf{RoI Localization.}
% First, we observe that all problematic local semantic edits share a common first step: localizing a Region of Interest (RoI) in the image according to the text instruction.  
All problematic local semantic edits share a common first step: localizing a Region of Interest (RoI) in the image for editing. 
Given an image \( I \) and an editing instruction \( T \), we first employ a Large Language Model (LLM) \cite{openai2024gpt4technicalreport} to locate the text RoI:
\begin{equation}
  \rho = M_{\mathrm{LLM}}(T), \label{eq:llm}
\end{equation}
where \( \rho \) represents the text RoI extracted by the LLM \( M_{\mathrm{LLM}}\). 
Subsequently, we achieve accurate localization of image RoI by:
\begin{equation}
  R = M_{\mathrm{seg}}(I, \rho), \label{eq:seg}
\end{equation}
% where \( M_{\mathrm{seg}} \) denotes the segmentation model \cite{yuan2025sa2vamarryingsam2llava} that returns the image RoI \( R \).
where \( R \) denotes the image RoI segmented by the segmentation model \( M_{\mathrm{seg}} \) \cite{yuan2025sa2vamarryingsam2llava}.

For add operation, the instruction may not specify a text RoI, or the specification may be ambiguous. In such cases, we first derive the overall layout of all candidate objects using the capability of segmentation models \cite{ren2024groundedsamassemblingopenworld,yuan2025sa2vamarryingsam2llava}, and then prompt the LLM to determine the appropriate image RoI based on \(T\).

\begin{figure}[t]
    \vspace{-0.3cm}
    \centering 
    \includegraphics[width=1.0\textwidth]{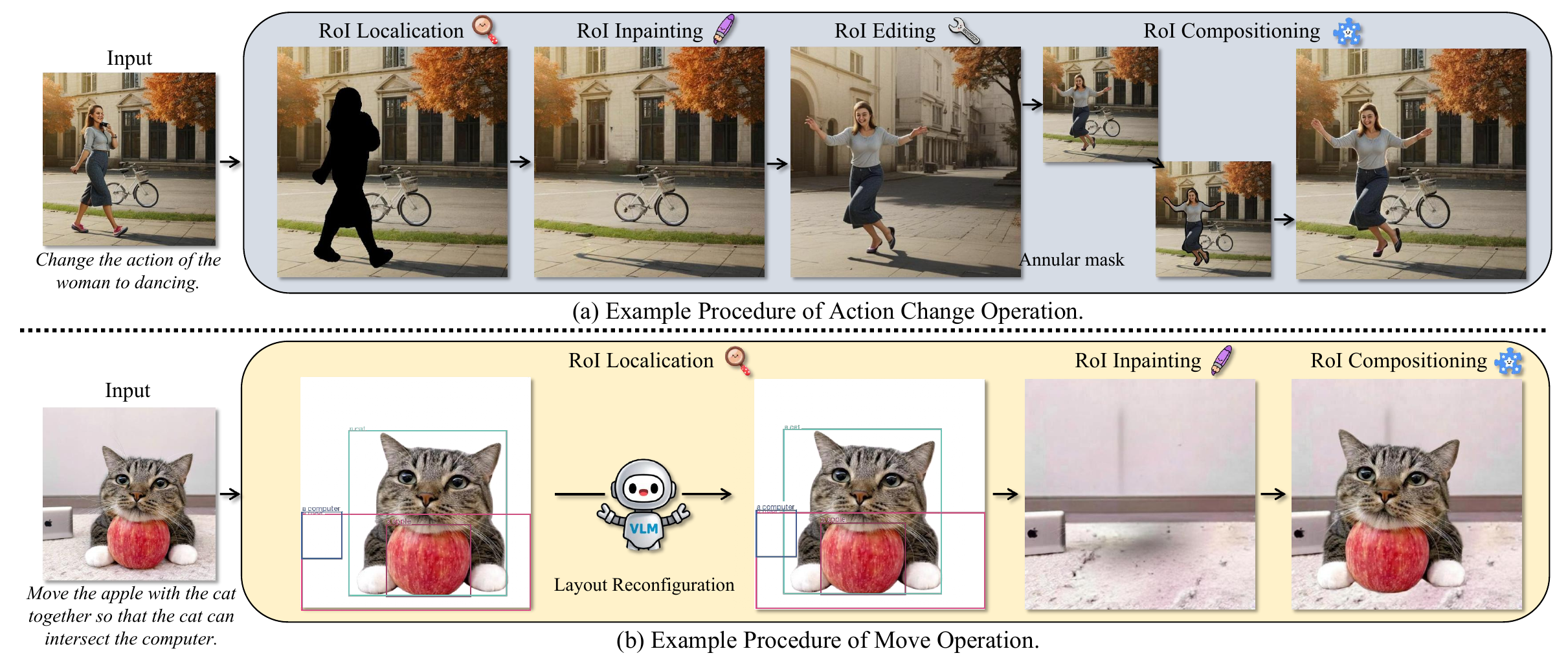}
    \caption{Example procedure. Figure (a) and Figure (b) illustrate the procedures of action change and movement respectively.}
    \vspace{-0.3cm}
    \label{fig:method} 
\end{figure}

Regarding move and resize, once the image RoI is obtained, we update the spatial layout of the image using an LLM \cite{openai2024gpt4technicalreport}. Specifically, we provide the LLM with a set of in-context examples that define our layout representation and demonstrate representative editing patterns \cite{lian2024llmgroundeddiffusionenhancingprompt}. Given the current layout \(L\) and the instruction \(T\), the LLM is prompted to produce a modified layout \(L_{\text{edit}}\), as formulated below:
\begin{equation}
\text{Tags} = M_{\mathrm{LLM}}(I),\quad L = M_{\mathrm{seg}}(\text{Tags}),\quad L_{\text{edit}} = M_{\mathrm{LLM}}(L, T).
\end{equation}

We then derive the geometric differences between \(L\) and \(L_{\text{edit}}\) and convert them into the corresponding affine transformations, consisting of translation, scaling, and reshaping, and apply it to \(R\) to update the spatial configuration, yielding the transformed mask \(R'\).

\noindent \textbf{RoI Inpainting.} Once the image RoI has been localized, we apply inpainting to seamlessly fill and complete the region. For additive and substitutive operations, which aim to introduce new objects, we employ a prompt-conditioned inpainting process to guide the generation of new content. Specifically, we first extract the semantic entity \(E\) from the instruction \(T\) via an LLM \cite{openai2024gpt4technicalreport}:
\begin{equation}
  E = M_{\mathrm{LLM}}(T), \label{eq:llm2}
\end{equation}
and then construct a composite prompt \(P\) in the form: \textit{``add \(E\) on the black region''}. 
For removal operations, which aim to eliminate existing content without introducing new semantics, we adopt a background-oriented infilling strategy, setting \(P\) as \textit{``fill in the hole of the image''}. The edited image \(I_{\text{edit}}\) is then generated by:
\begin{equation}
  I_{\text{edit}} = M_{\mathrm{inpaint}}\left( I \odot (1 - R), P \right), \label{eq:inpaint}
\end{equation}
where \(M_{\mathrm{inpaint}}\) denotes the inpainting model trained by us.

\noindent \textbf{RoI Editing.} When operations pertain to property change are performed, we use the trained attribute editing model \(M_{\mathrm{attr}}\) to perform edits in this stage to obtain $I_{\text{edit}}$:  
\begin{equation}
  I_{\text{edit}} = M_{\mathrm{attr}}\left( I , T \right). \label{eq:attr}
\end{equation}

\noindent \textbf{RoI Compositing.} To ensure seamless integration of the edited RoI with its surrounding context, we first construct an annular mask \(M_{\mathrm{ann}}\) by applying morphological dilation and erosion \cite{rivest1993morphological,said2021analysis} to the transformed RoI mask \(R'\):
\begin{equation}
  M_{\mathrm{ann}} = \mathrm{Dilate}(R',\,k_1)\;\setminus\;\mathrm{Erode}(R',\,k_2).
\end{equation}
Then, we employ a fusion network \(M_{\mathrm{fusion}}\), trained on ring-masked object boundaries, to refine the pre-composited image \(I_{\mathrm{prep}}\) using the generated annular mask. The final edited image is obtained as:

\begin{equation}
  I_{\text{edit}} = M_{\mathrm{fusion}}\left( I_{\mathrm{prep}} \odot (1 - M_{\mathrm{ann}}), P \right), 
\end{equation}
where \(P\) is set as ``inpaint the black-bordered region so that the object's edges blend smoothly with the background'' to guide seamless boundary blending.

\noindent \textbf{Global Transformation.} Like RoI editing, in the scenarios involving global transformation, we use the trained global transformation model \(M_{\mathrm{global}}\) to perform edits in this final stage to obtain $I_{\text{edit}}$.

\begin{figure}[t]
    \vspace{-0.3cm}
    \centering 
    \includegraphics[width=1.0\textwidth]{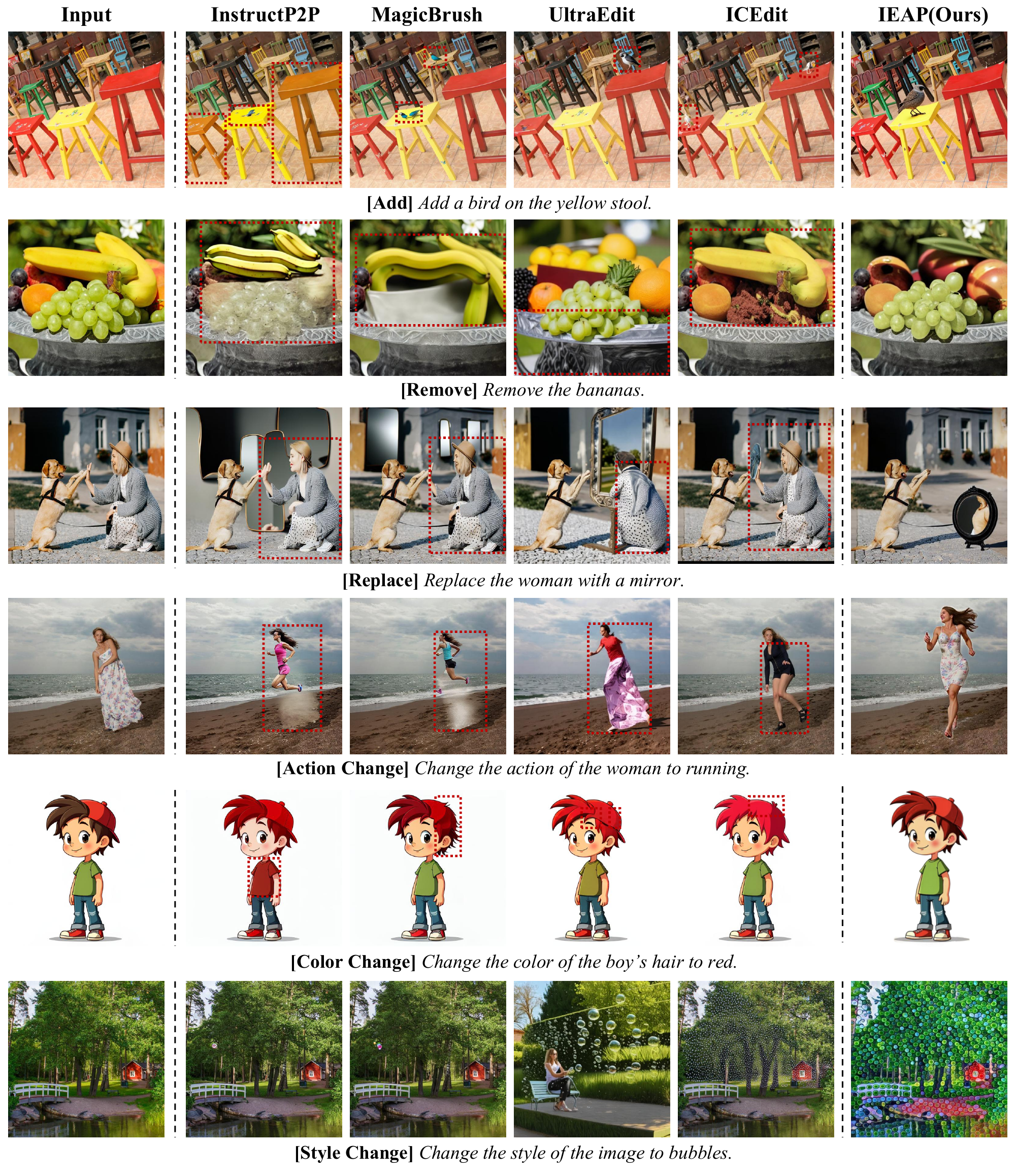}
    \vspace{-0.3cm}
    \caption{Comparison results of ours with baseline methods on representative editing cases. Others exhibit poor performance even on some common editing operations, while our approach demonstrates superior effectiveness across all operations.} 
    % Compared with others, our method demonstrates demonstrates superior performance in terms of instruction fidelity, entity consistency and background preservation.
    \label{fig:compare_results} 
    \vspace{-0.3cm}
\end{figure}

\section{Experiments}
\subsection{Experimental Settings}
\noindent \textbf{Training Settings.} We train four specialized models for RoI inpainting, RoI editing, RoI compositing, and global transformation respectively. 
All models are fine-tuned on FLUX.1-dev \cite{Flux2024} using LoRA \cite{hu2021loralowrankadaptationlarge}, with default settings for rank 128 and alpha 128. 
Training is conducted with a batch size of 1 and runs for 50,000 iterations each. We use the Prodigy optimizer \cite{mishchenko2023prodigy}, enabling safeguard warmup and bias correction, with a weight decay of 0.01.
The experiments are conducted on single NVIDIA H100 GPU (80GB).

\noindent \textbf{Dataset Setup.} For both the RoI editing and global transformation models, we sample from the relevant subsets of the AnyEdit \cite{yu2024anyedit} dataset and apply GPT-4o \cite{openai2024gpt4ocard} to filter the data of some types that have numerous noisy examples. 
To cover facial expression edits absent in AnyEdit, we integrate the CelebHQ-FM dataset \cite{decann2022comprehensivedatasetfacemanipulations}, which offers consistent identities and annotated expressions suitable for our instruction schema. 
The RoI inpainting and RoI compositing models are trained on samples from the ``add'', ``remove'' and ``replace'' splits of AnyEdit. For each sample, we first obtain the image RoI according to the editing instruction. In the RoI Inpainting training setup, we set the pixels within image RoI to black as input to train. For RoI Compositing, we set $k_1$ and $k_2$ as 3 in default to blackout the annular mask region of image RoI as input for training.  

\noindent \textbf{Evaluation Settings.}  We evaluate our method on two benchmarks: MagicBrush test set \cite{zhang2023magicbrush}, a widely used dataset spanning diverse editing types, and AnyEdit test set \cite{yu2024anyedit}, from which we select 16 instruction-based editing categories. 
For MagicBrush, we follow previous works \cite{zhang2023magicbrush,zhao2024ultraedit,fu2023guiding,sheynin2024emu} and report CLIPimg, CLIPout \cite{hessel2021clipscore}, $L_1$, and DINO \cite{caron2021emerging,oquab2023dinov2} scores to measure the similarity between the generated results and ground-truth images.
While for AnyEdit, where some categories lack reference captions required for calculating CLIPout, we instead leverage GPT-4o \cite{openai2024gpt4ocard} to assign ratings on a scale from 1 to 5 across three aspects: instruction faithfulness, semantic consistency, and aesthetic quality. The final quality score is computed as the average of these three dimensions.

We first compare our method with existing state-of-the-art open-source baselines, including InstructPix2Pix \cite{brooks2023instructpix2pix}, MagicBrush \cite{zhang2023magicbrush}, UltraEdit \cite{zhao2024ultraedit}, and ICEdit \cite{zhang2025incontexteditenablinginstructional}.
In addition, to demonstrate the competitiveness of our approach against powerful proprietary multimodal foundation models in complex image editing scenarios, we further make comparisons with SeedEdit (Doubao) \cite{shi2024seededit}, Gemini 2.0 Flash \cite{GoogleGemini2025}, and GPT-4o \cite{openai2024gpt4ocard}. 
\subsection{Comparisons with State of the Art.}
\label{stateoftheart}

\begin{table}[t]
\vspace{-0.3cm}
\centering
\small
\renewcommand\arraystretch{1}
\begin{tabular}{l|cccc|cccc}
\toprule
\multirow{2}{*}{Method} & \multicolumn{4}{c|}{MagicBrush test} & \multicolumn{4}{c}{AnyEdit test} \\
\cmidrule(r){2-9} 
 & CLIP$_{im}$ $\uparrow$ & CLIP$_{out}$ $\uparrow$ & L1 $\downarrow$ & DINO $\uparrow$ & CLIP$_{im}$ $\uparrow$ & L1 $\downarrow$ & DINO $\uparrow$ & GPT $\uparrow$ \\
\midrule
InstructPix2Pix  & 0.838 & 0.229 & 0.112 & 0.758 & 0.801 & \underline{0.110} & 0.765 & 3.83 \\ 
MagicBrush  & 0.886 & \underline{0.241} & 0.074 & 0.859 & 0.824 & 0.128 & 0.742 & 3.90 \\ 
UltraEdit  & 0.911 & 0.227 & 0.061 & 0.883 & 0.833 & 0.114 & \underline{0.772} & 3.93 \\ 
ICEdit & \underline{0.913} & 0.236 & \textbf{0.058} & \underline{0.885} & \underline{0.847} & \underline{0.110} & 0.765 & \underline{4.13} \\ 
\midrule
Ours & \textbf{0.922} & \textbf{0.247} & \underline{0.060} & \textbf{0.897} & \textbf{0.882} & \textbf{0.096} & \textbf{0.825} & \textbf{4.41}\\
\bottomrule
\end{tabular}
\vspace{+0.03cm}
\caption{Quantitative results on MagicBrush and AnyEdit test set.}
\label{tab: exp_metric}
\vspace{-0.25cm}
\end{table}

\begin{table}[t]
\vspace{-0.25cm}
\centering
\small
\renewcommand\arraystretch{1}
\setlength{\tabcolsep}{1pt}
\begin{tabular}{l|cccc|cccc|cccc}
\toprule
\multirow{2}{*}{Method} & \multicolumn{4}{c|}{Local Semantic Editing} & \multicolumn{4}{c|}{Local Attribute Editing} & \multicolumn{4}{c}{Overall Content Editing}\\
\cmidrule(r){2-6} \cmidrule(l){6-10}  \cmidrule(l){10-13} 
 & CLIP$_{im}$ $\uparrow$  & L1 $\downarrow$ & DINO $\uparrow$ & GPT $\uparrow$ & CLIP$_{im}$ $\uparrow$ & L1 $\downarrow$ & DINO $\uparrow$ & GPT $\uparrow$ & CLIP$_{im}$ $\uparrow$ & L1 $\downarrow$ & DINO $\uparrow$ & GPT $\uparrow$ \\
\midrule
InstructP2P  & 0.826 & 0.132 & 0.738 & 3.74 & 0.790 & 0.135 & 0.737 & 3.92 & \underline{0.766} & \underline{0.156} & \underline{0.642} & 3.91  \\ 
MagicBrush  & 0.860 & 0.106 & 0.796 & 3.90 & 0.809 & 0.117 & 0.762 & 4.21 & 0.763 & 0.187 & 0.616 & 3.99  \\ 
UltraEdit  & 0.867 & 0.095 & \underline{0.812} & 3.86 & 0.801 & \underline{0.092} & 0.793 & 3.94 & 0.754 & 0.201 & 0.611 & 4.41  \\ 
ICEdit & \underline{0.881} & \underline{0.088} & 0.810 & \underline{4.08} & \underline{0.825} & 0.095 & \underline{0.795} & \underline{4.06} & 0.759 & 0.188 & 0.603 & \underline{4.45}  \\ 
\midrule
Ours & \textbf{0.907} & \textbf{0.081} & \textbf{0.854} & \textbf{4.42} & \textbf{0.861} & \textbf{0.083} & \textbf{0.821} & \textbf{4.54} & \textbf{0.895} & \textbf{0.107} & \textbf{0.879} & \textbf{4.51} \\
\bottomrule
\end{tabular}
\vspace{+0.03cm}
\caption{Quantitative results on different types of editing operations.}
\label{tab: exp_metric_type}
\vspace{-0.5cm}
\end{table}

\noindent \textbf{Qualitative Comparisons.} Fig. \ref{fig:compare_results} shows the results of ours against other four methods \cite{brooks2023instructpix2pix,zhang2023magicbrush,zhao2024ultraedit,zhang2025incontexteditenablinginstructional} on some representative editing cases, where our method demonstrates comprehensive superiority over others in accurate instruction execution, structural consistency, and instance-level fidelity.

\noindent \textbf{Quantitative Comparisons.} Table \ref{tab: exp_metric} exhibits the quantitative comparison results of our method and other approaches \cite{brooks2023instructpix2pix,zhang2023magicbrush,zhao2024ultraedit,zhang2025incontexteditenablinginstructional} on MagicBrush test set \cite{zhang2023magicbrush} and AnyEdit test set \cite{yu2024anyedit}.
The results show that our method demonstrates state-of-the-art performance on both datasets.
On MagicBrush, our method achieves the best performance in terms of caption alignment, semantic consistency, and preservation of fine‐grained structural details. Although it incurs a marginal increase in pixel‐level deviation compared to the best \cite{zhang2025incontexteditenablinginstructional}, this is far outweighed by the substantial gains in perceptual quality and semantic fidelity. Furthermore, on AnyEdit, our approach yields significant and comprehensive improvements across all evaluation metrics, further highlighting its superiority over existing techniques.

To provide a more fine‐grained analysis of editing performance, we group a subset of the instruction‐based categories from the AnyEdit test set \cite{yu2024anyedit} into three macro‐tasks: local semantic editing, local attribute editing and overall semantic editing. For local attribute editing, we augment with some CelebHQ-FM \cite{decann2022comprehensivedatasetfacemanipulations} test images to evaluate facial expression changes.
The quantitave comparison results are shown in Tab. \ref{tab: exp_metric_type}, where our method consistently outperforms other candidates across all three task categories and evaluation metrics.

\begin{figure}[t]
    \vspace{-0.35cm}
    \centering 
    \includegraphics[width=1.0\textwidth]{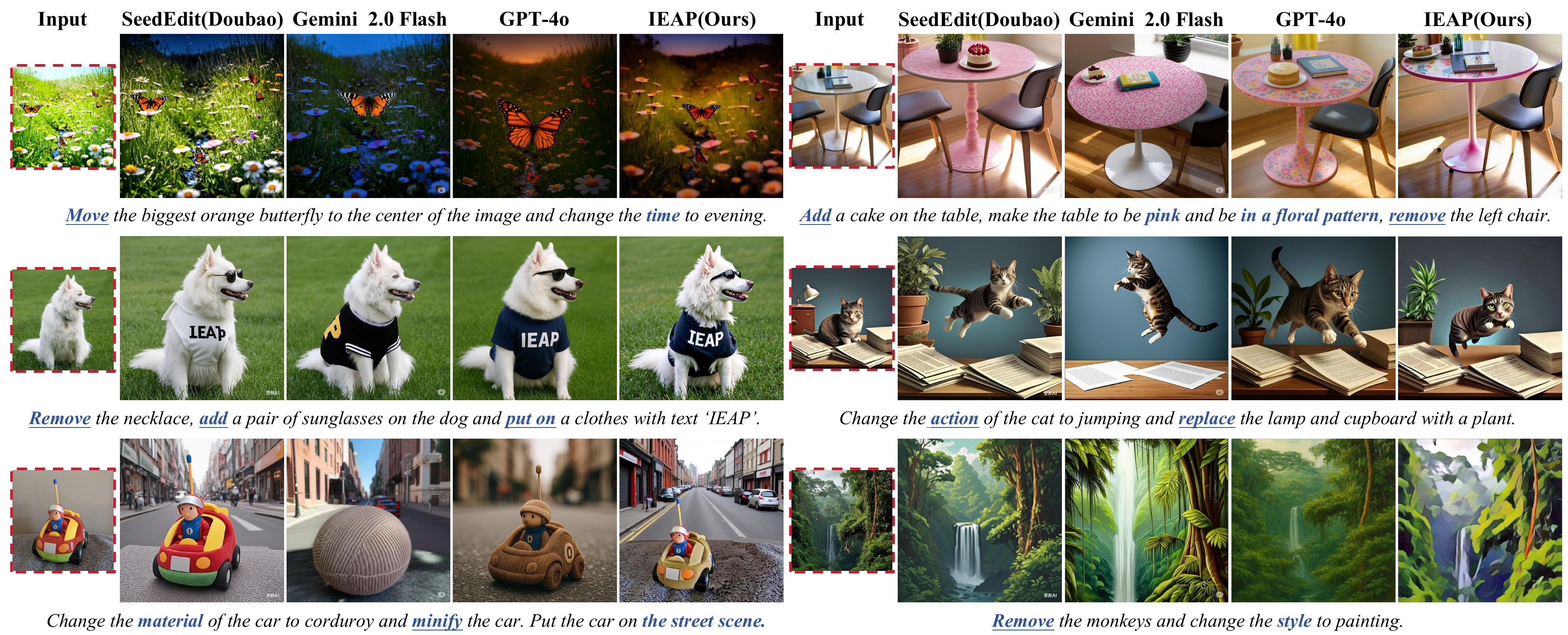}
    \caption{Comparisons on Complex Instructions with Leading Multimodal Models. Our method achieves comparable or even better edit completeness and pre-post consistency.}
    \vspace{-0.35cm}
    \label{fig:comm}
\end{figure}

\noindent \textbf{Comparisons with Cutting-Edge Multimodal Models.} To demonstrate the superiority of our reduction strategy on complex editing tasks, we also conduct comparative experiments against prominent closed-source multimodal models \cite{shi2024seededit,GoogleGemini2025,openai2024gpt4ocard}. As illustrated in Fig. \ref{fig:comm}, our method rivals, and in most cases surpasses the performance of these leading models on intricate scenarios requiring multiple sequential edits. Unlike competing approaches, which frequently omit specified instructions or introduce extraneous alterations unrelated to the editing directives, our framework faithfully executes each instruction while maintaining superior image consistency and instance preservation.

\subsection{Ablation Studies}
\begin{figure}[H]
\vspace{-0.4cm}
\centering
\begin{minipage}[ht]{0.62\textwidth}
    \centering
    \small
    \renewcommand\arraystretch{1}
    \setlength{\tabcolsep}{0.5pt}
    \begin{tabular}{lccccc}
    \toprule
    Settings & CLIP$_{im}$ $\uparrow$ & CLIP$_{out}$ $\uparrow$ & L1 $\downarrow$ & DINO $\uparrow$ & GPT $\uparrow$ \\
    \midrule
    w/o CoT \& Reduction & 0.873 & 0.241 & 0.117 & 0.795 & 4.10 \\
    \midrule
    w/o RoI Inpainting & 0.861 & 0.218 & 0.124 & 0.775 & 3.65 \\
    w/o RoI Editing & 0.900 & 0.244 & 0.088 & 0.843 & 4.23 \\
    w/o Layout Reconfiguration & 0.900 & 0.245 & 0.088 & 0.848 & 4.31 \\
    w/o Annular Mask Integration & 0.906 & \textbf{0.252} & 0.083 & \textbf{0.854} & 4.39 \\
    \midrule
    Full & \textbf{0.907} & \textbf{0.252} & \textbf{0.081} & \textbf{0.854} & \textbf{4.42} \\
    \bottomrule
    \end{tabular}
    \vspace{-0.2cm}
    \captionof{table}{Ablation results on AnyEdit local semantic editing test set.}
    \label{tab: ablation}
\end{minipage}
\vspace{-0.3cm}
\hfill
\begin{minipage}[ht]{0.33\textwidth}
    \centering
    \vspace{-0.2cm}
    \includegraphics[width=\linewidth]{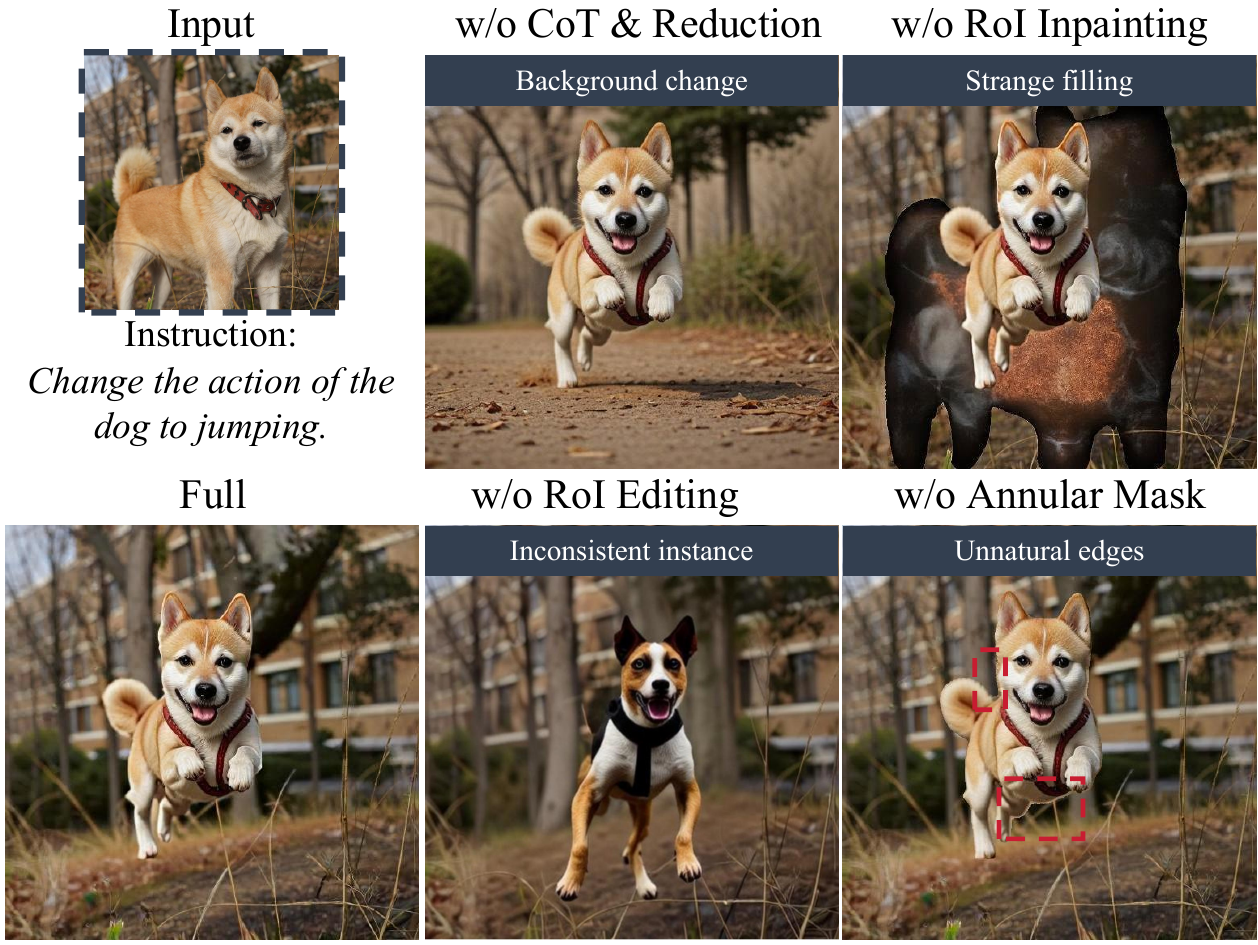}
    % \vspace{-em}
    \vspace{-0.63cm}
    \caption{Qualitative ablation of action change operation.}
    \label{fig:ablation_vis}
\end{minipage}
\vspace{-0.3cm}
\end{figure}
\noindent \textbf{Module-wise Ablation Studies. } To quantify the impact of each key component in our framework, we perform a series of ablation studies on the AnyEdit local semantic editing test set as we split in Sec. \ref{stateoftheart}.
As shown in Tab. \ref{tab: ablation}, we first substitute our CoT reasoning and reduction pipeline with end-to-end editing pipeline, resulting in a marked performance deterioration across all metrics. 
Next, we replace our specialized RoI inpainting and RoI editing models respectively with the generic inpainting model from \cite{tan2024ominicontrol}, which induces performance declines of varying degrees.
We then remove the LLM-guided layout reconfiguration and instead employing random layout modifications for relevant operations, which incurs a noticeable performance decline.
Finally, omitting the annular mask integration produces a modest drop, underscoring its role in precise boundary delineation.
Fig. \ref{fig:ablation_vis} exhibits the ablation results on an example of ``action change'', visually showcasing each module’s necessity.
Collectively, these ablation results confirm that each component in our pipeline contributes significantly in handling robust local semantic editing tasks requiring layout changes.

\section{Conclusions, Limitations and Future Work}
% In this paper, we propose Image Editing As Programs (IEAP), a unified framework for instruction-driven image editing built on DiT architecture. 
% Our comprehensive taxonomy and empirical study reveals a pronounced performance gap between structurally-consistent and structurally-inconsistent edits in existing DiT-based methods. 
% Fortunately, by introducing a small set of five atomic operations and leveraging CoT reasoning to compile free-form instructions into sequential programs, IEAP is capable of executing both simple and complex editing instructions. 
% Experiments demonstrate that IEAP outperforms state-of-the-art approaches in both structure-preserving and structure-altering scenarios, delivering superior accuracy and semantic fidelity, especially in complex, multi-step editing tasks.
In this paper, we propose Image Editing As Programs (IEAP), a unified DiT-based framework for instruction-driven image editing. By defining five atomic operations and using CoT reasoning to convert instructions into sequential programs, IEAP processes the ability to handle both simple and complex edits. Experiments demonstrate that IEAP outperforms state-of-the-art methods in both structure-preserving and structure-altering tasks, especially for complex edits.

Despite its strong overall performance, there are also some limitations. First, for complex shadow changes, our method sometimes leaves shadows inconsistent after compositing operations. Second, multiple editing iterations may induce progressive image quality decay.
Future work could focus on addressing these issues via physics-aware shadow modeling and diffusion-based quality restoration.

{\small
\bibliographystyle{plain}
\bibliography{main}
}
{
% % \small
% \bibliographystyle{IEEEtran}
% \bibliography{reference}
% }

%%%%%%%%%%%%%%%%%%%%%%%%%%%%%%%%%%%%%%%%%%%%%%%%%%%%%%%%%%%%

\appendix
\section*{Technical Appendices and Supplementary Material}

% \section{Technical Appendices and Supplementary Material}
% Technical appendices with additional results, figures, graphs and proofs may be submitted with the paper submission before the full submission deadline (see above), or as a separate PDF in the ZIP file below before the supplementary material deadline. There is no page limit for the technical appendices.

In this part, we provide additional algorithm illustration, implementation details, more comparison results, more visualization results, and more analysis and discussions of the proposed approach. 

\section{Algorithm Illustration}
To better elaborate the details of the proposed IEAP, we provide an algorithmic illustration for the whole pipeline in Alg. \ref{alg:ieap}.

\begin{algorithm}[ht]
\caption{IEAP: Image Editing As Programs}
\label{alg:ieap}
\textbf{Input:} 
\begin{itemize}[leftmargin=*]
  \item $I$: input image path
  \item $T$: original instruction 
  \item $\{\texttt{RoI\_Localization},\ \texttt{RoI\_Inpainting},\ \dots,\ \texttt{Global\_Transformation}\}$: editing primitives
  \item $\texttt{cot\_with\_gpt}(\cdot)$: CoT prompt to GPT–4o
  \item $\texttt{extract\_instructions}(\cdot)$: parse CoT output
  \item $\texttt{infer\_with\_DiT}(\texttt{op}, \cdot)$: invoke DiT for primitive \texttt{op}
  \item $\texttt{roi\_localization}(I, instr)$: returns mask for region of interest
  \item $\texttt{fusion}(I_1, I_2)$: blends two intermediate outputs
  \item $\texttt{layout\_change}(I, instr)$: compute geometric transform
\end{itemize}
\textbf{Output:} final edited image $I^*$  
\begin{algorithmic}[1]
\State $uri \leftarrow \texttt{encode\_image\_to\_datauri}(I)$
\State $(\mathcal{C}, \mathcal{T}) \leftarrow \texttt{cot\_with\_gpt}(uri, T)$ 
\Comment{Categories and instructions}
\State $I^{(0)} \leftarrow I$
\For{$i = 1$ to $|\mathcal{C}|$}
  \State $cat \leftarrow \mathcal{C}[i]$, \quad $instr \leftarrow \mathcal{T}[i]$
  \If{$cat\in \{\texttt{Add}, \texttt{Remove}, \texttt{Replace}\}$}
    \State $M \leftarrow \texttt{roi\_localization}(I^{(i-1)}, instr)$
    \State $I'\leftarrow\texttt{infer\_with\_DiT}(\texttt{RoI Inpainting}, M, instr)$
    \State $I^{(i)}\leftarrow I'$
  \ElsIf{$cat = \texttt{Action Change}$}
    \State $M \leftarrow \texttt{roi\_localization}(I^{(i-1)}, instr)$
    \State $I_{bg}\leftarrow \texttt{infer\_with\_DiT}(\texttt{RoI Inpainting}, M, instr)$
    \State $I_{act}\leftarrow \texttt{infer\_with\_DiT}(\texttt{RoI Editing}, I^{(i-1)}, instr)$
    \State $I^{(i)}\leftarrow \texttt{infer\_with\_DiT}(\texttt{RoI Compositing}, \texttt{fusion}(I_{bg},I_{act}), instr)$
  \ElsIf{$cat \in \{\texttt{Move}, \texttt{Resize}\}$}
    \State $M \leftarrow \texttt{roi\_localization}(I^{(i-1)}, instr)$
    \State $I_{bg}\leftarrow \texttt{infer\_with\_DiT}(\texttt{RoI Inpainting}, M, instr)$
    \State $I_{lc}\leftarrow \texttt{layout\_change}(I^{(i-1)}, instr)$
    \State $I^{(i)}\leftarrow \texttt{infer\_with\_DiT}(\texttt{RoI Compositing}, \texttt{fusion}(I_{bg},I_{lc}), instr)$
  \ElsIf{$cat \in \{\texttt{Appearance Change}, \texttt{Background Change},$\\\quad$\texttt{Color Change}, \texttt{Material Change}, \texttt{Expression Change}\}$}
    \State $I^{(i)}\leftarrow \texttt{infer\_with\_DiT}(\texttt{RoI Editing}, I^{(i-1)}, instr)$
  \ElsIf{$cat \in \{\texttt{Tone Transfer}, \texttt{Style Change}\}$}
    \State $I^{(i)}\leftarrow \texttt{infer\_with\_DiT}(\texttt{Global Transformation}, I^{(i-1)}, instr)$
  \Else
    \State \textbf{raise} ValueError(``Invalid category: ''${cat}$'')
  \EndIf
\EndFor
\State \Return $I^{(|\mathcal{C}|)}$
\end{algorithmic}
\end{algorithm}

\section{Implementation Details}
In this section, we present the prompts employed to leverage a VLM for CoT reasoning over complex instructions, providing further details on the layout-adjustment prompts. 

Below are the detailed prompts used to invoke the VLM for the CoT process on complex instructions:

\begin{center}
\fcolorbox{black}{white!0}{\parbox{1\linewidth}{
Now you are an expert in image editing. Based on the given single image, what atomic image editing instructions should be if the user wants to \{instruction\}? Let's think step by step.\\ 
Atomic instructions include 13 categories as follows:\\
- Add: Introduce a new object, person, or element into the image, e.g.: add a car on the road\\
- Remove: Eliminate an existing object or element from the image, e.g.: remove the sofa in the image\\
- Color Change: Modify the color of a specific object, e.g.: change the color of the shoes to blue\\
- Material Change: Alter the surface material or texture of an object, e.g.: change the material of the sign like stone\\
- Action Change: Modify the pose or action of an instance, e.g.: change the action of the boy to raising hands\\
- Expression Change: Adjust the facial expression, e.g.: change the expression to smiling\\
- Replace: Substitute one object in the image with a different object, e.g.: replace the coffee with an apple\\
- Background Change: Change the background scene to another, e.g.: change the background into forest\\
- Appearance Change: Modify visual attributes such as patterns or accessories, e.g.: make the cup have a floral pattern\\
- Move: Change the spatial position of an object within the image, e.g.: move the plane to the left\\
- Resize: Adjust the scale or size of an object, e.g.: enlarge the clock\\
- Tone Transfer: Change the global atmosphere or lighting conditions, e.g.: change the weather to foggy, change the time to spring\\
- Style Change: Modify the entire image to adopt a different visual style, e.g.: make the style of the image to cartoon\\
Respond *only* with a numbered list.  
Each line must begin with the category in square brackets, then the instruction. Please strictly follow the atomic categories.  
The operation (what) and the target (to what) are crystal clear.  
Do not split replace to add and remove.  
Always place [Tone Transfer] and [Style Change] instructions at the end of the list.  \\
For example:  \\
1. [Add] add a car on the road  \\
2. [Color Change] change the color of the shoes to blue  \\
3. [Move] move the lamp to the left  \\
Do not include any extra text, explanations, JSON or markdown, just the list.
}}
\end{center}

Below are the detailed prompts used to adjust the layout of move and resize operations:

\begin{center}
\fcolorbox{black}{white!0}{\parbox{1\linewidth}{
You are an intelligent bounding box editor. I will provide you with the current bounding boxes and the editing instruction. Your task is to generate the new bounding boxes after editing. Let's think step by step.\\
The images are of size 512x512. The top-left corner has coordinate [0, 0]. The bottom-right corner has coordinnate [512, 512]. 
The bounding boxes should not overlap or go beyond the image boundaries. 
Each bounding box should be in the format of (object name, [top-left x coordinate, top-left y coordinate, bottom-right x coordinate, bottom-right y coordinate]). \\
Do not add new objects or delete any object provided in the bounding boxes. Do not change the size or the shape of any object unless the instruction requires so.\\
Please consider the semantic information of the layout. 
When resizing, keep the bottom-left corner fixed by default. When swaping locations, change according to the center point. \\
If needed, you can make reasonable guesses. Please refer to the examples below:\\
Input bounding boxes: [("bed", [50, 300, 450, 450]), ("pillow", [200, 200, 300, 230])]\\
Editing instruction: Move the pillow to the left side of the bed.\\
Output bounding boxes: [("bed", [50, 300, 450, 450]), ("pillow", [70, 270, 170, 300])]
}}
\end{center}

\begin{center}
\fcolorbox{black}{white!0}{\parbox{1\linewidth}{   
% Input bounding boxes: [('a green car', [21, 281, 232, 440]), ('a blue truck', [269, 283, 478, 443]), ('a red air balloon', [66, 8, 211, 143]), ('a bird', [296, 42, 439, 142])]\\
% Editing instruction: Move the car to the right.\\
% Output bounding boxes: [('a green car', [81, 281, 292, 440]), ('a blue truck', [269, 283, 478, 443]), ('a red air balloon', [66, 8, 211, 143]), ('a bird', [296, 42, 439, 142])]\\
% Input bounding boxes: [("sofa", [100, 300, 400, 400]), ("dog", [150, 250, 250, 300])]\\
Editing instruction: 
                    % Enlarge the dog.\\
                    % Output bounding boxes: [("sofa", [100, 300, 400, 400]), ("dog", [150, 225, 300, 300])]\\                  
                    % Input bounding boxes: [("chair", [100, 350, 200, 450]), ("lamp", [300, 200, 360, 300])]\\
                    % Editing instruction: Swap the location of the chair and the lamp.\\
                    % Output bounding boxes: [("chair", [280, 200, 380, 300]), ("lamp", [120, 350, 180, 450])]\\
                    Input bounding boxes: [('a car', [21, 281, 232, 440])]\\
                    Editing instruction: Move the car to the right.\\
                    Output bounding boxes: [('a car', [121, 281, 332, 440])]\\               
                    Input bounding boxes: [("dog", [150, 250, 250, 300])]\\
                    Editing instruction: Enlarge the dog.\\
                    Output bounding boxes: [("dog", [150, 225, 300, 300])]\\
                    Input bounding boxes: [("chair", [100, 350, 200, 450]), ("lamp", [300, 200, 360, 300])]\\
                    Editing instruction: Swap the location of the chair and the lamp.\\
                    Output bounding boxes: [("chair", [280, 200, 380, 300]), ("lamp", [120, 350, 180, 450])]\\
                    Now, the current bounding boxes is \{bbox\}, the instruction is \{instruction\}. 
                    
}}
\end{center}

Below are the detailed prompts used to adjust the layout of add operations:

\begin{center}
\fcolorbox{black}{white!0}{\parbox{1\linewidth}{
You are an intelligent bounding box editor. I will provide you with the current bounding boxes and an add editing instruction. 
                    Your task is to determine the new bounding box of the added object. Let's think step by step.\\
                    The images are of size 512x512. The top-left corner has coordinate [0, 0]. The bottom-right corner has coordinnate [512, 512]. \\
                    The bounding boxes should not go beyond the image boundaries. 
                    The new box must be at least as large as needed to encompass the object.
                    Each bounding box should be in the format of (object name, [top-left x coordinate, top-left y coordinate, bottom-right x coordinate, bottom-right y coordinate]). 
                    Do not delete any object provided in the bounding boxes. 
                    Please consider the semantic information of the layout, preserve semantic relations. \\
                    If needed, you can make reasonable guesses. Please refer to the examples below:\\
                    % Input bounding boxes: [('a green car', [21, 281, 232, 440])]\\
                    % Editing instruction: Add a bird on the green car.\\
                    % Output bounding boxes: [('a bird', [80, 150, 180, 281]), ('a green car', [21, 281, 232, 440])]\\
                    % Input bounding boxes: [('stool', [300, 350, 380, 450])]\\
                    % Editing instruction: Add a cat to the left of the stool.\\
                    % Output bounding boxes: [('a cat', [180, 300, 300, 450])]\\
                    % Input bounding boxes: [('the white cat', [200, 300, 320, 420])]\\
                    % Editing instruction: Add a hat on the white cat.\\
                    % Output bounding boxes: [('the white hat', [200, 260, 320, 310]), ('cat', [200, 300, 320, 420])]\\
                    Input bounding boxes: [('a green car', [21, 281, 232, 440])]\\
                    Editing instruction: Add a bird on the green car.\\
                    Output bounding boxes: [('a bird', [80, 150, 180, 281])]\\
                    Input bounding boxes: [('stool', [300, 350, 380, 450])]\\
                    Editing instruction: Add a cat to the left of the stool.\\
                    Output bounding boxes: [('a cat', [180, 250, 300, 450])]\\
                    Here are some examples to illustrate appropriate overlapping for better visual effects:\\
                    Input bounding boxes: [('the white cat', [200, 300, 320, 420])]\\
                    Editing instruction: Add a hat on the white cat.\\
                    Output bounding boxes: [('a hat', [200, 150, 320, 330])]\\
                    Now, the current bounding boxes is \{bbox\}, the instruction is \{instruction\}.
}}
\end{center}

\section{More Quantitative Results}
\begin{table}[H]
\centering
\small
\renewcommand\arraystretch{1}
\setlength{\tabcolsep}{4.0pt}
\begin{tabular}{l|cccccccc}
\toprule
Method & CLIP$_{im} \uparrow$ & CLIP$_{out} \uparrow$ & L1 $\downarrow$ & DINO $\uparrow$ & GPT$_{IF} \uparrow$ & GPT$_{FC} \uparrow$ & GPT$_{AQ} \uparrow$ & GPT$_{avg} \uparrow$ \\
\midrule
InstructPix2Pix  & 0.847 &  0.264  & 0.092 & 0.829 & 4.50 & 4.40 & 4.26 & 4.39 \\ 
MagicBrush       & 0.889 &  \underline{0.277}  & 0.068 & 0.892 & \underline{4.66} & 4.76 & \underline{4.62} & 4.68 \\ 
UltraEdit        & 0.897 & 0.274 & \textbf{0.056} & 0.909 & 3.36 & 4.24 & 4.22 & 3.94 \\ 
ICEdit           & \underline{0.925} & \underline{0.277}  & 0.057 & \underline{0.915} & 4.60 & \underline{4.80} & \textbf{4.76} & \textbf{4.72} \\ 
\midrule
IEAP(Ours)       & \textbf{0.928} & \textbf{0.278} & \textbf{0.056} & \textbf{0.917} & \textbf{4.68} & \textbf{4.84} & 4.60 & \underline{4.71} \\
\bottomrule
\end{tabular}
\vspace{+0.1cm}
\caption{Quantitative comparison results on AnyEdit Add test set.}
\label{tab:d1}
\vspace{-0.75cm}
\end{table}

\begin{table}[H]
\centering
\small
\renewcommand\arraystretch{1}
\setlength{\tabcolsep}{4.0pt}
\begin{tabular}{l|cccccccc}
\toprule
Method & CLIP$_{im} \uparrow$ & CLIP$_{out} \uparrow$ & L1 $\downarrow$ & DINO $\uparrow$ & GPT$_{IF}\uparrow$ & GPT$_{FC}\uparrow$ & GPT$_{AQ}\uparrow$ & GPT$_{avg}\uparrow$ \\
\midrule
InstructPix2Pix  & 0.800 & 0.202 & 0.108 & 0.721 & 2.74 & 3.42 & 3.20 & 3.12 \\ 
MagicBrush       & 0.853 & \underline{0.211} & 0.083 & 0.800 & 3.08 & 3.60 & 3.18 & 3.29 \\ 
UltraEdit        & 0.846 & \underline{0.211} & 0.066 & 0.802 & 2.50 & 3.54 & 3.44 & 3.16 \\ 
ICEdit           & \underline{0.895} & 0.212 & \textbf{0.054} & \underline{0.875} & \underline{4.06} & \textbf{4.48} & \textbf{4.32} & \textbf{4.29} \\ 
\midrule
IEAP(Ours)       & \textbf{0.916} & \textbf{0.230} & \underline{0.057} & \textbf{0.886} & \textbf{4.18} & \underline{3.88} & \underline{3.66} & \underline{3.91} \\
\bottomrule
\end{tabular}
\vspace{+0.1cm}
\caption{Quantitative comparison results on AnyEdit Remove test set.}
\label{tab:d2}
\vspace{-0.75cm}
\end{table}

\begin{table}[H]
\centering
\small
\renewcommand\arraystretch{1}
\setlength{\tabcolsep}{4.0pt}
\begin{tabular}{l|cccccccc}
\toprule
Method & CLIP$_{im} \uparrow$ & CLIP$_{out} \uparrow$ & L1 $\downarrow$ & DINO $\uparrow$ & GPT$_{IF}\uparrow$ & GPT$_{FC}\uparrow$ & GPT$_{AQ}\uparrow$ & GPT$_{avg}\uparrow$ \\
\midrule
InstructPix2Pix  & 0.766 & 0.234 & 0.179 & 0.588 & 3.72 & 3.68 & 3.80 & 3.73 \\ 
MagicBrush       & \underline{0.806} & \underline{0.248} & 0.148 & \underline{0.671} & \underline{4.52} & \underline{4.48} & 4.38 & \underline{4.46} \\ 
UltraEdit        & 0.779 & 0.242 & 0.142 & 0.621 & 3.80 & 4.40 & \underline{4.40} & 4.20 \\ 
ICEdit           & 0.797 & 0.228 & \underline{0.128} & 0.614 & 3.68 & 4.02 & 4.04 & 3.91 \\ 
\midrule
IEAP(Ours)       & \textbf{0.866} & \textbf{0.252} & \textbf{0.099} & \textbf{0.701} & \textbf{4.68} & \textbf{4.68} & \textbf{4.48} & \textbf{4.61} \\
\bottomrule
\end{tabular}
\vspace{+0.1cm}
\caption{Quantitative comparison results on AnyEdit Replace test set.}
\label{tab:d3}
\vspace{-0.75cm}
\end{table}

\begin{table}[H]
\centering
\small
\renewcommand\arraystretch{1}
\setlength{\tabcolsep}{4.0pt}
\begin{tabular}{l|cccccccc}
\toprule
Method & CLIP$_{im} \uparrow$ & CLIP$_{out} \uparrow$ & L1 $\downarrow$ & DINO $\uparrow$ & GPT$_{IF}\uparrow$ & GPT$_{FC}\uparrow$ & GPT$_{AQ}\uparrow$ & GPT$_{avg}\uparrow$ \\
\midrule
InstructPix2Pix  & 0.829 & 0.254 & 0.164 & 0.774 & \underline{3.46} & 3.84 & 3.58 & 3.63 \\ 
MagicBrush       & 0.831 & \underline{0.266} & 0.156 & \underline{0.784} & 2.96 & \underline{4.28} & \underline{4.28} & \underline{3.84} \\ 
UltraEdit        & \underline{0.847} & 0.259 & 0.157 & 0.781 & 2.92 & 4.22 & 4.24 & 3.79 \\ 
ICEdit           & 0.827 & 0.255 & \textbf{0.152} & 0.745 & 2.68 & 4.04 & 4.04 & 3.59 \\ 
\midrule
IEAP(Ours)       & \textbf{0.848} & \textbf{0.267} &\underline{0.154} & \textbf{0.798} & \textbf{4.66} & \textbf{4.86} & \textbf{4.68} & \textbf{4.73} \\
\bottomrule
\end{tabular}
\vspace{+0.1cm}
\caption{Quantitative comparison results on AnyEdit Action Change test set.}
\label{tab:d4}
\vspace{-0.75cm}
\end{table}

\begin{table}[H]
\centering
\small
\renewcommand\arraystretch{1}
\setlength{\tabcolsep}{4.0pt}
\begin{tabular}{l|cccccccc}
\toprule
Method & CLIP$_{im} \uparrow$ & CLIP$_{out} \uparrow$ & L1 $\downarrow$ & DINO $\uparrow$ & GPT$_{IF}\uparrow$ & GPT$_{FC}\uparrow$ & GPT$_{AQ}\uparrow$ & GPT$_{avg}\uparrow$ \\
\midrule
InstructPix2Pix  & 0.881 & \underline{0.219} & 0.127 & 0.771 & \underline{3.82} & \underline{4.44} & 4.36 & \underline{4.21} \\ 
MagicBrush       & 0.902 & \underline{0.219} & 0.088 & 0.828 & 2.94 & 3.94 & 3.90 & 3.59 \\ 
UltraEdit        & 0.923 & 0.211 & 0.074 & 0.867 & 3.48 & 4.40 & \textbf{4.40} & 4.09 \\ 
ICEdit           & \underline{0.944} & 0.213 & \underline{0.063} & \underline{0.868} & 3.28 & \textbf{4.64} & 4.30 & 4.07 \\ 
\midrule
IEAP(Ours)       & \textbf{0.963} & \textbf{0.223} & \textbf{0.058} & \textbf{0.903} & \textbf{3.88} & \underline{4.44} & \underline{4.38} & \textbf{4.23} \\
\bottomrule
\end{tabular}
\vspace{+0.1cm}
\caption{Quantitative comparison results on AnyEdit Relation test set.}
\label{tab:d5}
\vspace{-0.75cm}
\end{table}

\begin{table}[H]
\centering
\small
\renewcommand\arraystretch{1}
\setlength{\tabcolsep}{4.0pt}
\begin{tabular}{l|cccccccc}
\toprule
Method & CLIP$_{im} \uparrow$ & CLIP$_{out} \uparrow$ & L1 $\downarrow$ & DINO $\uparrow$ & GPT$_{IF}\uparrow$ & GPT$_{FC}\uparrow$ & GPT$_{AQ}\uparrow$ & GPT$_{avg}\uparrow$ \\
\midrule
InstructPix2Pix  & 0.831 & 0.241 & 0.124 & 0.746 & 2.94 & 3.56 & 3.62 & 3.37 \\ 
MagicBrush       & 0.875 & 0.258 & 0.094 & 0.802 & 2.80 & 3.88 & 4.00 & 3.56 \\ 
UltraEdit        & \underline{0.908} & \underline{0.262} & \underline{0.073} & \underline{0.889} & \underline{3.22} & \textbf{4.38} & \textbf{4.38} & \underline{4.00} \\ 
ICEdit           & 0.895 & 0.253 & 0.074 & 0.841 & 3.14 & 4.28 & 4.26 & 3.89 \\ 
\midrule
IEAP(Ours)       & \textbf{0.923} & \textbf{0.263} & \textbf{0.066} & \textbf{0.921} & \textbf{4.38} & \underline{4.32} & \underline{4.28} & \textbf{4.32} \\
\bottomrule
\end{tabular}
\vspace{+0.1cm}
\caption{Quantitative comparison results on AnyEdit Resize test set.}
\label{tab:d6}
\vspace{-0.75cm}
\end{table}

\begin{table}[H]
\centering
\small
\renewcommand\arraystretch{1}
\setlength{\tabcolsep}{4.0pt}
\begin{tabular}{l|cccccccc}
\toprule
Method & CLIP$_{im} \uparrow$ & CLIP$_{out} \uparrow$ & L1 $\downarrow$ & DINO $\uparrow$ & GPT$_{IF}\uparrow$ & GPT$_{FC}\uparrow$ & GPT$_{AQ}\uparrow$ & GPT$_{avg}\uparrow$ \\
\midrule
InstructPix2Pix  & 0.815 & 0.280 & 0.139 & 0.744 & 3.60 & 4.08 & 3.92 & 3.87 \\ 
MagicBrush       & 0.852 & \textbf{0.294} & 0.094 & 0.815 & 3.96 & 4.32 & 3.98 & 4.09 \\ 
UltraEdit        & \underline{0.857} & 0.277 & \textbf{0.068} & \textbf{0.845} & \underline{4.04} & \underline{4.62} & \underline{4.42} & \underline{4.36} \\ 
ICEdit           & 0.847 & 0.273 & 0.085 & 0.808 & \underline{4.04} & 4.42 & 4.16 & 4.21 \\ 
\midrule
IEAP(Ours)       & \textbf{0.886} & \underline{0.285} & \underline{0.082} & \underline{0.833} & \textbf{4.06} & \textbf{4.72} & \textbf{4.80} & \textbf{4.53} \\
\bottomrule
\end{tabular}
\vspace{+0.1cm}
\caption{Quantitative comparison results on AnyEdit Appearance test set.}
\label{tab:d7}
\vspace{-0.75cm}
\end{table}

\begin{table}[H]
\centering
\small
\renewcommand\arraystretch{1}
\setlength{\tabcolsep}{4.0pt}
\begin{tabular}{l|cccccccc}
\toprule
Method & CLIP$_{im} \uparrow$ & CLIP$_{out} \uparrow$ & L1 $\downarrow$ & DINO $\uparrow$ & GPT$_{IF}\uparrow$ & GPT$_{FC}\uparrow$ & GPT$_{AQ}\uparrow$ & GPT$_{avg}\uparrow$ \\
\midrule
InstructPix2Pix  & 0.725 & 0.224 & 0.216 & 0.582 & 3.40 & 3.60 & 3.44 & 3.48 \\ 
MagicBrush       & 0.746 & 0.230 & 0.228 & 0.567 & \underline{4.58} & \underline{4.38} & \underline{4.46} & \underline{4.47} \\ 
UltraEdit        & 0.796 & \textbf{0.257} & 0.169 & 0.747 & 3.48 & 4.36 & 3.14 & 3.66 \\ 
ICEdit           & \underline{0.799} & 0.241 & \underline{0.166} & \underline{0.757} & 3.04 & 4.16 & 3.88 & 3.69 \\ 
\midrule
IEAP(Ours)       & \textbf{0.801} & \underline{0.243} & \textbf{0.165} & \textbf{0.759} & \textbf{4.74} & \textbf{4.68} & \textbf{4.70} & \textbf{4.71} \\
\bottomrule
\end{tabular}
\vspace{+0.1cm}
\caption{Quantitative comparison results on AnyEdit Background Change test set.}
\label{tab:d8}
\vspace{-0.75cm}
\end{table}

\begin{table}[H]
\centering
\small
\renewcommand\arraystretch{1}
\setlength{\tabcolsep}{4.0pt}
\begin{tabular}{l|ccccccccc}
\toprule
Method & CLIP$_{im} \uparrow$ & CLIP$_{out} \uparrow$ & L1 $\downarrow$ & DINO $\uparrow$ & GPT$_{IF}\uparrow$ & GPT$_{FC}\uparrow$ & GPT$_{AQ}\uparrow$ & GPT$_{avg}\uparrow$ \\
\midrule
InstructPix2Pix  & 0.886 & 0.279 & 0.120 & \textbf{0.876} & 3.60 & 4.40 & 4.00 & 4.00 \\ 
MagicBrush       & \underline{0.898} & \textbf{0.282} & 0.087 & 0.869 & 4.20 & \textbf{4.82} & 4.62 & 4.55 \\ 
UltraEdit        & 0.890 & \underline{0.280} & \underline{0.065} & 0.87 & 3.80 & 4.40 & 4.20 & 4.13 \\ 
ICEdit           & 0.896 & 0.278 & 0.073 & 0.849 & \textbf{4.72} & \underline{4.80} & \underline{4.64} & \textbf{4.72} \\ 
\midrule
IEAP(Ours)       & \textbf{0.911} & 0.276 & \textbf{0.059} & \textbf{0.876} & \underline{4.62} & 4.72 & \textbf{4.78} & \underline{4.71} \\
\bottomrule
\end{tabular}
\vspace{+0.1cm}
\caption{Quantitative comparison results on AnyEdit Color Change test set.}
\label{tab:d9}
\vspace{-0.75cm}
\end{table}

\begin{table}[H]
\centering
\small
\renewcommand\arraystretch{1}
\setlength{\tabcolsep}{6.8pt}
\begin{tabular}{l|ccccccc}
\toprule
Method & CLIP$_{im} \uparrow$ & L1 $\downarrow$ & DINO $\uparrow$ & GPT$_{IF}\uparrow$ & GPT$_{FC}\uparrow$ & GPT$_{AQ} \uparrow$ & GPT$_{avg}\uparrow$ \\
\midrule
InstructPix2Pix  & 0.776 & 0.068 & 0.936 & 3.74 & \underline{4.60} & \underline{4.30} & \underline{4.21} \\ 
MagicBrush       & 0.770 & \underline{0.064} & 0.940 & \underline{3.86} & 4.48 & 4.18 & 4.17 \\ 
UltraEdit        & 0.699 & 0.073 & 0.907 & 3.14 & 4.10 & 3.80 & 3.68 \\ 
ICEdit           & \underline{0.796} & 0.065 & \underline{0.943} & 3.16 & \underline{4.60} & \underline{4.30} & 4.02 \\ 
\midrule
IEAP(Ours)       & \textbf{0.882} & \textbf{0.052} & \textbf{0.945} & \textbf{4.34} & \textbf{4.72} & \textbf{4.50} & \textbf{4.52} \\
\bottomrule
\end{tabular}
\vspace{+0.1cm}
\caption{Quantitative comparison results on Expression test set.}
\label{tab:d10}
\vspace{-0.75cm}
\end{table}

\begin{table}[H]
\centering
\small
\renewcommand\arraystretch{1}
\setlength{\tabcolsep}{6.8pt}
\begin{tabular}{l|ccccccc}
\toprule
Method & CLIP$_{im} \uparrow$  & L1 $\downarrow$ & DINO $\uparrow$ & GPT$_{IF}\uparrow$ & GPT$_{FC}\uparrow$ & GPT$_{AQ}\uparrow$ & GPT$_{avg}\uparrow$ \\
\midrule
InstructPix2Pix  & 0.746 & 0.130 & 0.549 & \underline{4.00} & 4.18 & \underline{4.04} & \underline{4.07} \\ 
MagicBrush       & 0.778 & 0.110 & \underline{0.621} & 3.36 & 4.06 & 3.84 & 3.75 \\ 
UltraEdit        & 0.765 & \underline{0.086} & 0.598 & 3.34 & \underline{4.28} & \underline{4.04} & 3.89 \\ 
ICEdit           & \underline{0.787} & \underline{0.086} & 0.616 & 3.48 & 3.92 & 3.58 & 3.66 \\ 
\midrule
IEAP(Ours)       & \textbf{0.826} & \textbf{0.055} & \textbf{0.696} & \textbf{4.08} & \textbf{4.48} & \textbf{4.18} & \textbf{4.25} \\
\bottomrule
\end{tabular}
\vspace{+0.1cm}
\caption{Quantitative comparison results on Material Change test set.}
\label{tab:d11}
\vspace{-0.75cm}
\end{table}

\begin{table}[H]
\centering
\small
\renewcommand\arraystretch{1}
\setlength{\tabcolsep}{6.8pt}
\begin{tabular}{l|ccccccc}
\toprule
Method & CLIP$_{im} \uparrow$ & L1 $\downarrow$ & DINO $\uparrow$ & GPT$_{IF}\uparrow$ & GPT$_{FC}\uparrow$ & GPT$_{AQ}\uparrow$ & GPT$_{avg}\uparrow$ \\
\midrule
InstructPix2Pix  & \underline{0.710} & 0.212 & 0.463 & 3.56 & 4.32 & 3.94 & 3.94 \\ 
MagicBrush       & 0.692 & 0.214 & 0.440 & 3.12 & 4.64 & 4.00 & 3.92 \\ 
UltraEdit        & 0.703 & \underline{0.201} & \underline{0.467} & 4.02 & \underline{4.8} & \textbf{4.62} & \underline{4.48} \\ 
ICEdit           & 0.706 & 0.219 & 0.458 & \underline{4.04} & \textbf{4.82} & 4.36 & 4.41 \\ 
\midrule
IEAP(Ours)       & \textbf{0.922} & \textbf{0.097} & \textbf{0.915} & \textbf{4.44} & 4.64 & \underline{4.44} & \textbf{4.51} \\
\bottomrule
\end{tabular}
\vspace{+0.1cm}
\caption{Quantitative comparison results on AnyEdit Style Change test set.}
\label{tab:d12}
\vspace{-0.75cm}
\end{table}

\begin{table}[H]
\centering
\small
\renewcommand\arraystretch{1}
\setlength{\tabcolsep}{4.0pt}
\begin{tabular}{l|cccccccc}
\toprule
Method & CLIP$_{im} \uparrow$ & CLIP$_{out} \uparrow$ & L1 $\downarrow$ & DINO $\uparrow$ & GPT$_{IF}\uparrow$ & GPT$_{FC}\uparrow$ & GPT$_{AQ}\uparrow$ & GPT$_{avg}\uparrow$ \\
\midrule
InstructPix2Pix  & 0.822 & 0.260 & \textbf{0.100} & \underline{0.821} & 3.72 & 4.48 & 3.92 & 4.04 \\ 
MagicBrush       & \underline{0.834} & 0.266 & 0.159 & 0.791 & 3.56 & \underline{4.64} & 3.98 & 4.06 \\ 
UltraEdit        & 0.804 & \textbf{0.268} & 0.201 & 0.767 & \underline{4.12} & 4.62 & 4.26 & 4.33 \\ 
ICEdit           & 0.812 & 0.260 & 0.157 & 0.748 & 4.06 & \textbf{4.88} & \textbf{4.56} & \underline{4.50} \\ 
\midrule
IEAP(Ours)       & \textbf{0.868} & \textbf{0.268} & \underline{0.116} & \textbf{0.843} & \textbf{4.44} & \underline{4.64} & \underline{4.44} & \textbf{4.51} \\
\bottomrule
\end{tabular}
\vspace{+0.1cm}
\caption{Quantitative comparison results on AnyEdit Tone Transfer test set.}
\label{tab:d13}
\vspace{-0.75cm}
\end{table}

\begin{table}[H]
\centering
\small
\renewcommand\arraystretch{1}
\setlength{\tabcolsep}{6.8pt}
\begin{tabular}{l|ccccccc}
\toprule
Method & CLIP$_{im} \uparrow$ & L1 $\downarrow$ & DINO $\uparrow$ & GPT$_{IF}\uparrow$ & GPT$_{FC}\uparrow$ & GPT$_{AQ}\uparrow$ & GPT$_{avg}\uparrow$ \\
\midrule
InstructPix2Pix  & 0.815 & 0.134 & 0.647 & \underline{3.40} & 4.04 & \textbf{4.80} & \underline{4.08} \\ 
MagicBrush       & 0.835 & 0.081 & 0.697 & 1.82 & 3.56 & 3.50 & 2.96 \\ 
UltraEdit        & 0.833 & 0.066 & 0.756 & 2.58 & 4.02 & 4.02 & 3.54 \\ 
ICEdit           & \underline{0.906} & \textbf{0.042} & \textbf{0.842} & 2.98 & \underline{4.40} & 3.40 & 3.59 \\ 
\midrule
IEAP(Ours)       & \textbf{0.908} & \underline{0.056} & \underline{0.794} & \textbf{3.42} & \textbf{4.48} & \underline{4.46} & \textbf{4.12} \\
\bottomrule
\end{tabular}
\vspace{+0.1cm}
\caption{Quantitative comparison results on AnyEdit Counting test set.}
\label{tab:d14}
\vspace{-0.75cm}
\end{table}

\begin{table}[H]
\centering
\small
\renewcommand\arraystretch{1}
\setlength{\tabcolsep}{6.8pt}
\begin{tabular}{l|ccccccc}
\toprule
Method & CLIP$_{im} \uparrow$ & L1 $\downarrow$ & DINO $\uparrow$ & GPT$_{IF}\uparrow$ & GPT$_{FC}\uparrow$ & GPT$_{AQ}\uparrow$ & GPT$_{avg}\uparrow$ \\
\midrule
InstructPix2Pix  & 0.773 & 0.208 & 0.581 & 3.46 & 4.18 & 4.08 & 3.91 \\ 
MagicBrush       & 0.806 & 0.174 & 0.631 & 2.98 & 3.88 & 4.04 & 3.63 \\ 
UltraEdit        & \underline{0.825} & \textbf{0.167} & \textbf{0.669} & 2.82 & \underline{4.38} & \underline{4.38} & 3.86 \\ 
ICEdit           & 0.806 & 0.171 & 0.629 & \underline{3.56} & 4.16 & 4.06 & \underline{3.93} \\ 
\midrule
IEAP(Ours)       & \textbf{0.833} & \underline{0.169} & \underline{0.662} & \textbf{3.88} & \textbf{4.44} & \textbf{4.52} & \textbf{4.28} \\
\bottomrule
\end{tabular}
\vspace{+0.1cm}
\caption{Quantitative comparison results on AnyEdit Implicit Change test set.}
\label{tab:d15}
\vspace{-0.75cm}
\end{table}

\begin{table}[H]
\centering
\small
\renewcommand\arraystretch{1}
\setlength{\tabcolsep}{6.8pt}
\begin{tabular}{l|ccccccc}
\toprule
Method & CLIP$_{im} \uparrow$ & L1 $\downarrow$ & DINO $\uparrow$ & GPT$_{IF}\uparrow$ & GPT$_{FC}\uparrow$ & GPT$_{AQ}\uparrow$ & GPT$_{avg}\uparrow$ \\
\midrule
InstructPix2Pix  & 0.887 & 0.111 & 0.858 & \textbf{4.30} & \underline{4.50} & 4.30 & \textbf{4.37} \\ 
MagicBrush       & 0.900 & 0.100 & 0.874 & 4.12 & 4.36 & \textbf{4.54} & 4.34 \\ 
UltraEdit        & \underline{0.922} & \textbf{0.077} & \underline{0.911} & 3.24 & 4.4 & 4.36 & 4.00 \\ 
ICEdit           & 0.898 & \underline{0.079} & 0.864 & 4.16 & 4.46 & 4.20 & 4.27 \\ 
\midrule
IEAP(Ours)       & \textbf{0.938} & 0.084 & \textbf{0.925} & \underline{4.18} & \textbf{4.56} & \underline{4.38} & \textbf{4.37} \\
\bottomrule
\end{tabular}
\vspace{+0.1cm}
\caption{Quantitative comparison results on AnyEdit Move test set.}
\label{tab:d16}
\vspace{-0.75cm}
\end{table}

\begin{table}[H]
\centering
\small
\renewcommand\arraystretch{1}
\setlength{\tabcolsep}{4.0pt} 
\begin{tabular}{l|cccccccc}
\toprule
Method & CLIP$_{im} \uparrow$ & CLIP$_{out} \uparrow$ & L1 $\downarrow$ & DINO $\uparrow$ & GPT$_{IF}\uparrow$ & GPT$_{FC}\uparrow$ & GPT$_{AQ}\uparrow$ & GPT$_{avg}\uparrow$ \\
\midrule
InstructPix2Pix  & 0.688 & 0.243 & 0.189 & 0.742 & 1.04 & 4.38 & 3.92 & 3.11 \\ 
MagicBrush       & 0.680 & 0.255 & 0.156 & 0.786 & 1.02 & \underline{4.48} & 4.10 & 3.20 \\ 
UltraEdit        & 0.732 & 0.279 & \textbf{0.147} & \textbf{0.843} & 1.96 & 4.46 & 3.98 & 3.47 \\ 
ICEdit           & \textbf{0.810} & \textbf{0.289} & \underline{0.155} & \underline{0.811} & \textbf{4.18} & 4.42 & \textbf{4.68} & \textbf{4.43} \\ 
\midrule
IEAP(Ours)       & \underline{0.788} & \underline{0.285} & 0.162 & 0.786 & \underline{3.96} & \textbf{4.58} & \underline{4.06} & \underline{4.20} \\
\bottomrule
\end{tabular}
\vspace{+0.1cm}
\caption{Quantitative comparison results on AnyEdit Textual Change test set.}
\label{tab:d17}
\vspace{-0.75cm}
\end{table}

\section{More Visualization Results}
In this section, we provide more visualization results, as shown below:
\begin{figure}[ht]
    \centering
    \includegraphics[width=1\textwidth]{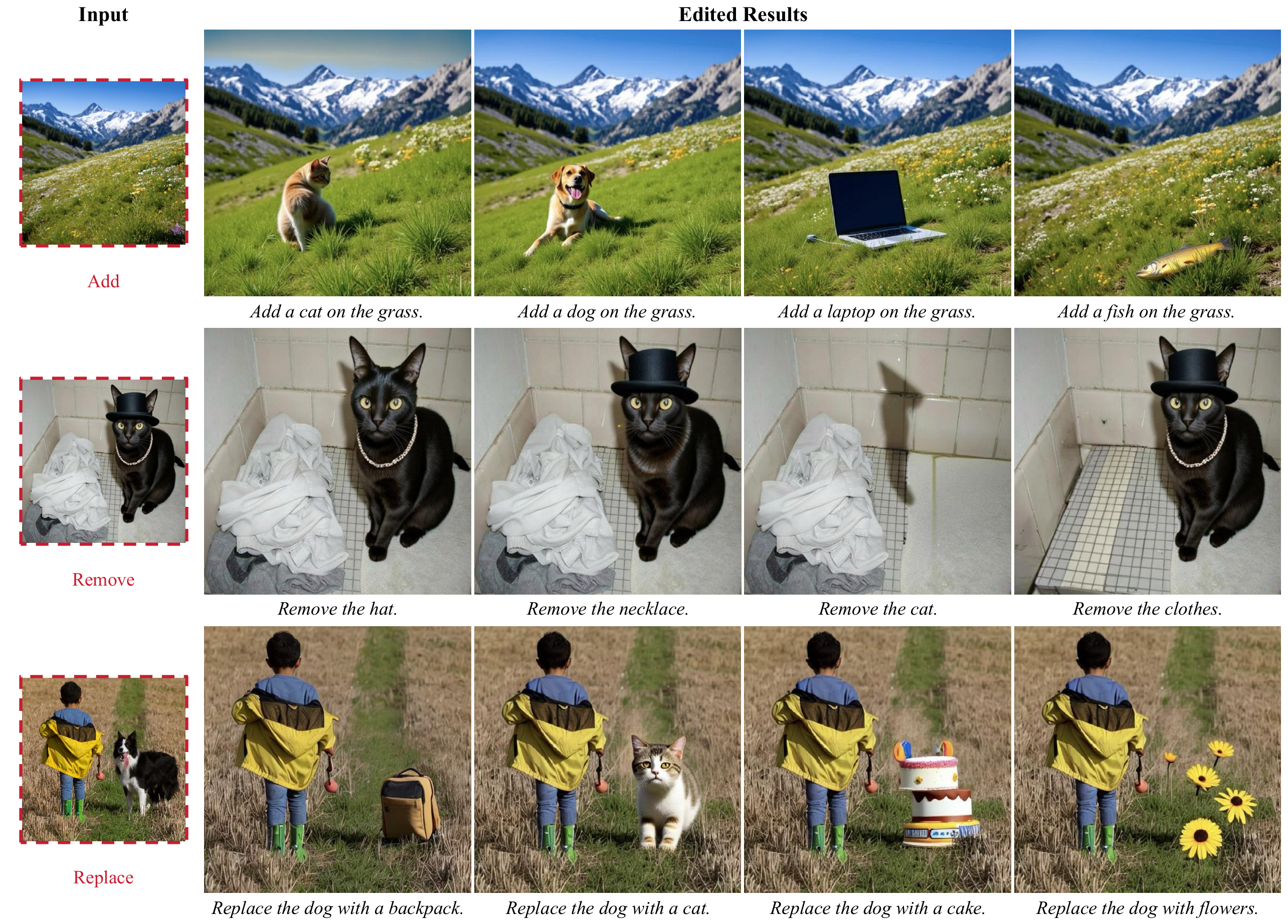} 
    \vspace{-0.35cm}
    \caption{More Visualization Results.}
    \label{fig:add1}
    \vspace{-0.75cm}
\end{figure}

\begin{figure}[ht]
    \centering
    \includegraphics[width=1\textwidth]{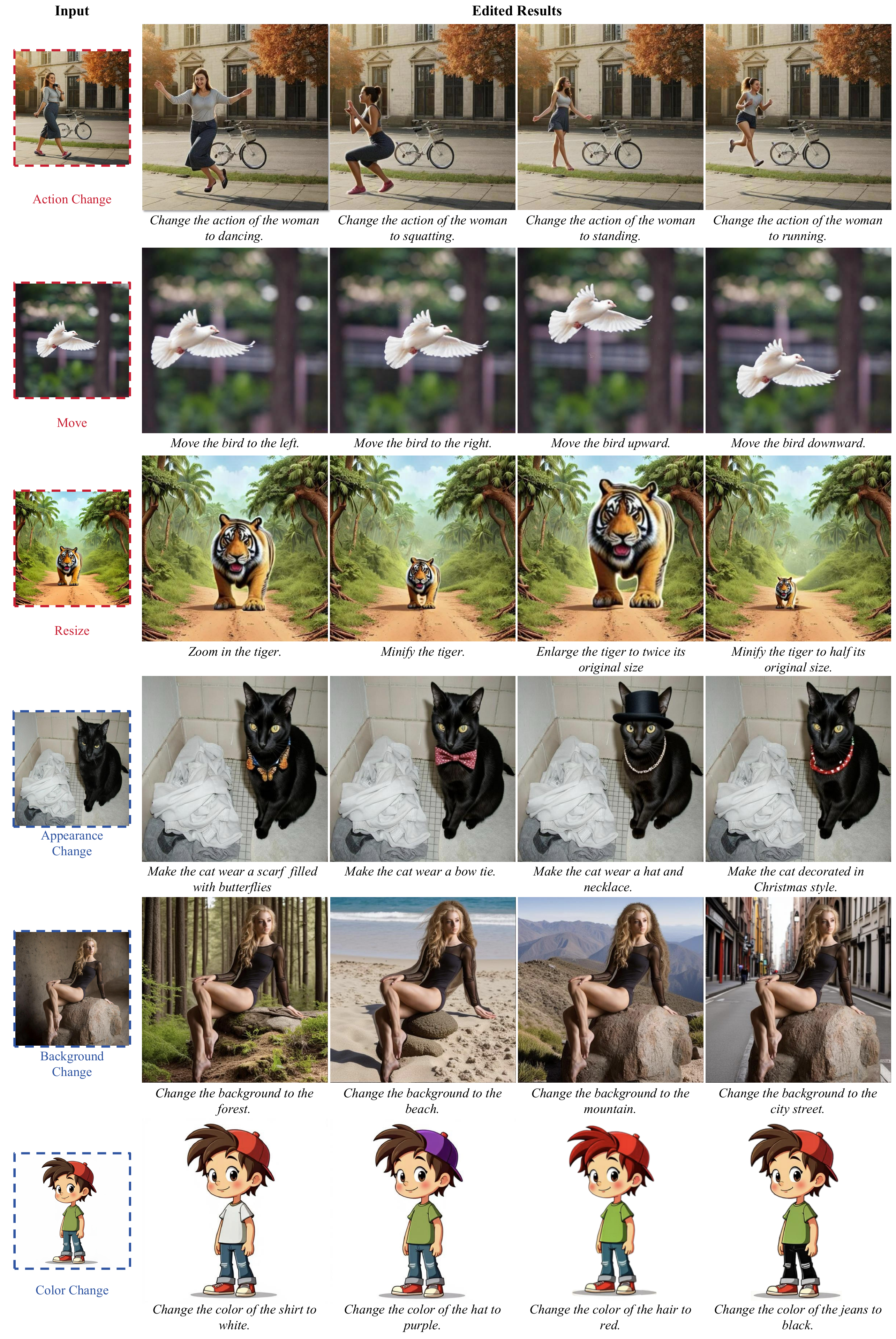} 
    \vspace{-0.35cm}
    \caption{More Visualization Results.}
    \label{fig:add2}
    \vspace{-0.35cm}
\end{figure}

\begin{figure}[ht]
    \centering
    \includegraphics[width=1\textwidth]{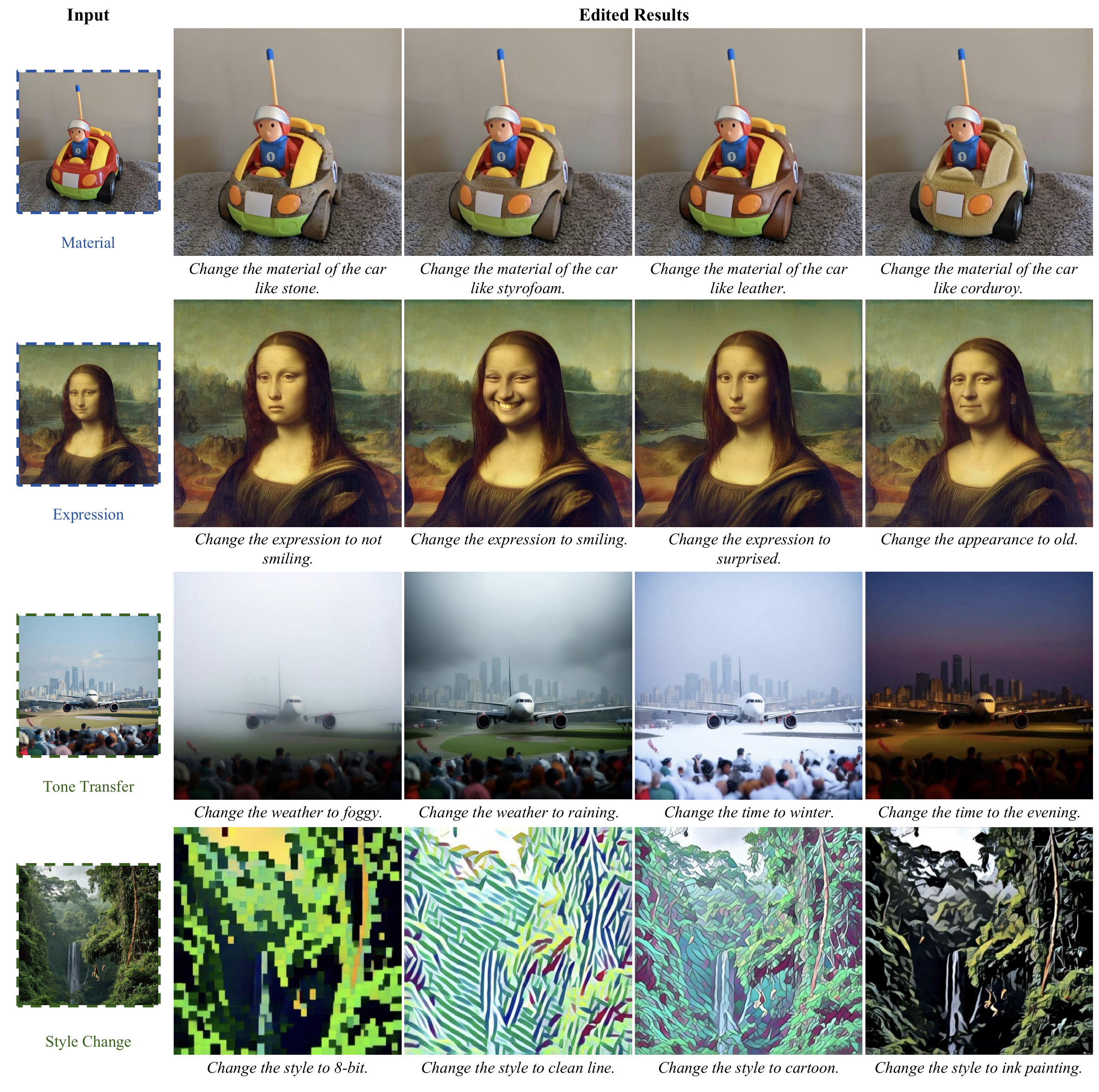} 
    \vspace{-0.35cm}
    \caption{More Visualization Results.}
    \label{fig:add3}
    \vspace{-0.95cm}
\end{figure}

\begin{figure}[ht]
    \centering
    \includegraphics[width=1\textwidth]{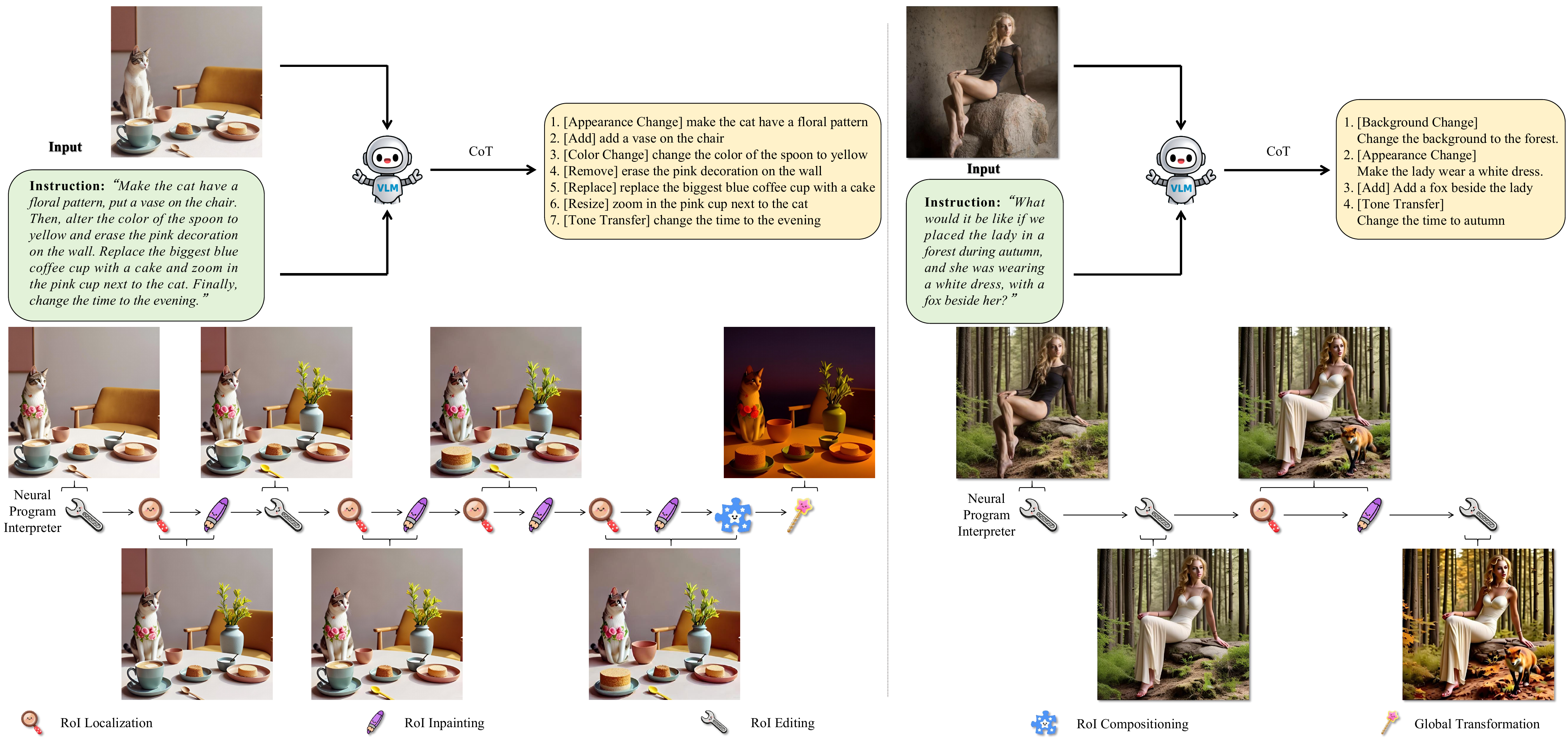} 
    \vspace{-0.35cm}
    \caption{More Detailed Visualization Processes of the pipeline.}
    \label{fig:add4}
    \vspace{-0.35cm}
\end{figure}

\FloatBarrier

\section{Analysis and Discussions}
\subsection{Runtime Performance Analysis}
We evaluate the time required for each atomic operation of IEAP on a single NVIDIA H100 GPU. Empirical measurements indicate that the RoI Localization stage requires approximately \SI{3}{\second} to \SI{5}{\second} per operation. Other editing primitives, including RoI Inpainting, RoI Editing, RoI Compositing, and Global Transformation, each consumes roughly \SI{7}{\second} to \SI{9}{\second} per operation. 

Consequently, a complete multi-step edit involving \(k\) atomic operations exhibits a total latency of
\[
T_\text{total} \;=\; \sum_{i=1}^{k} T_i
\quad\text{with}\quad
T_i =
\begin{cases}
\SIrange{3}{5}{\second}, & \text{if operation}_i = \text{RoI Localization},\\
\SIrange{7}{9}{\second}, & \text{otherwise}.
\end{cases}
\]
While this per-operation cost precludes real-time interactivity, it remains acceptable for batch-oriented workflows in digital content creation, scientific visualization, and other offline editing scenarios.

\subsection{Limitations and Future Work}
\noindent \textbf{Limitations.} Despite its strengths, IEAP exhibits several limitations in handling dynamic scenes and complex physical interactions. First, the RoI compositing may introduce geometric distortions or texture discontinuities when editing highly dynamic or non-rigid content, such as motion‐blurred instances, and fluid or smoke effects. For example, in the task of “changing the cat’s action to jumping,” in Fig. \ref{fig:comm}, the rapid motion of fur can produce blurred regions that fail to blend naturally with the background. 
Second,  RoI compositing struggles to simulate physically consistent lighting effects in scenes with reflective or refractive surfaces, sometimes resulting in mismatched shadow directions and illumination conflicts between edited objects and their environments. For example, in the task of “change the action of the woman to dancing,” in Fig. \ref{fig:method}, the shadows before and after editing remain the same, but the action of the woman has changed, so it is unnatural. Third, the DiT-based architecture and multi-stage atomic operations incur substantial inference latency for \SI{5}{\second} to \SI{9}{\second} per operation on a single H100 GPU, precluding real‐time interactivity in applications such as AR/VR. Finally, the requirement for high‐memory GPUs like NVIDIA H100 (80 GB) limits reproducibility for resource-constrained researchers, and multi‐iteration editing can exacerbate image quality degradation over successive operations.
  
\noindent \textbf{Future Work.} As for future work, several avenues may be pursued to overcome the identified limitations. 
To begin with, physics‐aware compositing techniques and motion‐compensated inpainting could be explored to better accommodate dynamic blur and fluid effects, thereby ensuring seamless integration of non‐rigid edits. Meanwhile, differentiable lighting models or neural rendering modules may be incorporated to enforce global illumination consistency, particularly in reflective and refractive contexts. On the performance front, model distillation, operation fusion, and sparse attention strategies could be investigated to reduce per‐operation latency and facilitate interactive editing. 
To enhance accessibility, memory optimization and support for smaller‐footprint architectures amenable to commodity GPUs may be implemented. Moreover, iterative refinement and error‐correction mechanisms may be developed to mitigate quality degradation over successive editing steps. 
Furthermore, beyond still‐image editing, an extension to video‐based complex instruction editing could be considered, where temporal coherence and motion consistency present additional challenges and opportunities for dynamic, multi‐step visual manipulation.  

\subsection{Societal Impacts and Ethical Safeguards}
\noindent \textbf{Positive Societal Impacts.} The proposed IEAP framework introduces a modular and interpretable approach to complex image editing, which holds significant potential to benefit a range of creative and technical domains. By decomposing high-level visual instructions into atomic operations, IEAP enables users to perform multi-step edits with enhanced precision and control. This capability is particularly valuable in digital content creation, advertising, and education, where fine-grained manipulation of visual content is often required. For example, IEAP’s ability to support structurally inconsistent modifications can streamline visual storytelling workflows or facilitate the generation of accurate scientific visualizations for publications and teaching materials. Furthermore, its potential extensions to fields such as medical imaging by enabling localized enhancement of diagnostic visuals, and accessibility technology by generating descriptive visual representations for users with visual impairments, demonstrate the framework’s broader societal utility and interdisciplinary relevance.

\noindent \textbf{Negative Societal Impacts and Ethical Safeguards.}
Despite its benefits, IEAP’s high-fidelity editing capabilities also introduce ethical risks, particularly in the domains of misinformation and privacy. The framework’s precision in altering visual content could be misused for the creation of deepfakes or manipulated images intended for disinformation, identity falsification, or reputational harm. Operations such as “Remove” or “Replace” could be exploited to tamper with sensitive or private imagery, potentially infringing on individual rights. 

To address these concerns, the development and deployment of IEAP adhere to strict ethical standards. Specifically, safeguards include the implementation of data filtering pipelines, such as the use of GPT-4o-filtered subsets of AnyEdit and the compliance-oriented CelebHQ-FM dataset, to reduce harmful biases and content. Additionally, the modular nature of IEAP facilitates transparency and traceability in the editing process, supporting future content provenance systems designed to detect and flag manipulated media. All these safeguards jointly contribute to ongoing efforts in AI safety and accountability.

\clearpage

\end{document}